\documentclass[preprint,12pt]{elsarticle}

\usepackage[margin=1in]{geometry}
\usepackage[utf8]{inputenc}
\usepackage{times}
\usepackage{amssymb, amsmath, amsbsy, mathtools}
\usepackage{makecell}  
\usepackage{graphicx}
\usepackage{float}
\usepackage{multirow}
\usepackage{xcolor}
\usepackage[export]{adjustbox}

\numberwithin{equation}{section}

\begin{document}

\begin{frontmatter}
    
\title{A nonlocal physics-informed deep learning framework using the peridynamic differential operator}

\author[MIT]{Ehsan Haghighat} 
\address[MIT]{Massachusetts Institute of Technology, Cambridge, MA}
\author[UA]{Ali Can Bekar}
\author[UA]{Erdogan Madenci}
\address[UA]{University of Arizona, Tucson, AZ}
\author[MIT]{Ruben Juanes}

\date{May 2020}


\begin{abstract}
    The Physics-Informed Neural Network (PINN) framework introduced recently incorporates physics into deep learning, and offers a promising avenue for the solution of partial differential equations (PDEs) as well as identification of the equation parameters. The performance of existing PINN approaches, however, may degrade in the presence of sharp gradients, as a result of the inability of the network to capture the solution behavior globally. We posit that this shortcoming may be remedied by introducing long-range (nonlocal) interactions into the network's input, in addition to the short-range (local) space and time variables. Following this ansatz, here we develop a \emph{nonlocal} PINN approach using the Peridynamic Differential Operator (PDDO)---a numerical method which incorporates long-range interactions and removes spatial derivatives in the governing equations. Because the PDDO functions can be readily incorporated in the neural network architecture, the nonlocality does not degrade the performance of modern deep-learning algorithms.  We apply nonlocal PDDO-PINN to the solution and identification of material parameters in solid mechanics and, specifically, to elastoplastic deformation in a domain subjected to indentation by a rigid punch, for which the mixed displacement--traction boundary condition leads to localized deformation and sharp gradients in the solution. We document the superior behavior of nonlocal PINN with respect to local PINN in both solution accuracy and parameter inference, illustrating its potential for simulation and discovery of partial differential equations whose solution develops sharp gradients.
\end{abstract}


\end{frontmatter}

\section{Introduction}\label{sec:introduction}

Deep learning has emerged as a powerful approach to computing-enabled knowledge in many fields \cite{Goodfellow2016}, such as image processing and classification \cite{bishop2006pattern,krizhevsky2012imagenet,Lecun2015}, search and recommender systems \cite{jannach2010recommender, zhang2019deep}, speech recognition \cite{graves2013speech}, autonomous driving \cite{bojarski2016end}, and healthcare \cite{miotto2018deep}. The particular needs of each application, collectively, have led to many different neural-network architectures, including Deep Neural Networks (DNN), Convolutional NNs (CNN), Recurrent NNs(RNN) and its variants including Long~Short-Term Memory RNNs (LSTM). Some of these frameworks have also been employed for data-driven modeling in computational mechanics \cite{brunton_kutz_2019}, including fluid mechanics and turbulent flow modeling \cite{brenner2019perspective}, solid mechanics and constitutive modeling \cite{ghaboussi1998new, kirchdoerfer2016data, haghighat2020deep}, and earthquake prediction and detection \cite{DeVries2018, Kong2018}. These efforts have resulted in the availability of open-source deep-learning platforms, including Theano \cite{bergstra2010theano}, Tensorflow \cite{abadi2016tensorflow}, and PyTorch \cite{paszke2019pytorch}. These software packages are highly efficient and ready to use on different platforms, from mobile devices to massively parallel cloud-based clusters, features that can be inherited in the development of tools for physics-informed deep learning \cite{haghighat2020sciann}.

Of particular interest to us are recent applications of deep learning in computational science and engineering, concerning the solution and discovery (identification) of partial differential equations describing various physical systems \cite{han2018solving,bar2019learning, rudy2019data, Raissi2019, Champion22445, raissi2020hidden}. Among these applications, a specific framework called Physics-Informed Neural Networks (PINN) \cite{Raissi2019} enables the construction of the solution space using feed-forward neural networks with space and time variables as the network's input. The governing equations are enforced in the loss function using automatic differentiation \cite{baydin2017automatic}. It is a framework that permits solving partial differential equations (PDEs) and conducting parameter identification (inversion) from data.  Multiple variations of this framework exist, such as Variational PINNs \cite{kharazmi2019variational} and Parareal PINNs \cite{meng2019ppinn}, which have been used for physics-informed learning of the Burgers equation, the Navier--Stokes equations, and the Schr\"{o}dinger equation.

Recently, PINN has been applied for inversion and discovery in solid mechanics \cite{haghighat2020deep}. While the method provides accurate and robust reconstructions and parameter estimates when the solution is smooth, the performance degrades in the presence of sharp gradients in the strain or stress fields. The emergence of near-discontinuities in the solution can occur for several reasons, including shear-band localization, crack propagation, and the presence of ``mixed'' displacement--traction boundary conditions. In the latter case, the point at which the boundary condition changes type often gives rise to stress concentration or even a stress singularity. In these cases, existing PINN approaches are much less accurate as a result of the inability of the network to capture the solution behavior globally. We posit that this shortcoming may be remedied by introducing long-range (nonlocal) interactions into the network's input, in addition to the short-range (local) space and time variables.

Here, we propose to use the Peridynamic Differential Operator (PDDO) \cite{madenci2016peridynamic,madenci2019peridynamic} to construct nonlocal neural networks with long-range interactions. Peridynamics, a nonlocal theory, was first introduced as an alternative to the local classical continuum mechanics to incorporate long-range interactions and to remove spatial derivatives in the governing equations \cite{silling2000reformulation,silling2005meshfree,silling2007peridynamic,silling2008convergence}. It has been shown to be well suited to model crack initiation and propagation \cite{madenci2014peridynamic}. It has also been shown that the peridynamic governing equations can be derived by replacing the local spatial derivatives in the Navier displacement equilibrium equations with their nonlocal representation using PDDO \cite{madenci2016peridynamic, madenci2017numerical, madenci2019peridynamic}. PDDO has an analytical form in terms of spatial integrals for a point with a symmetric interaction domain or support region. The PD functions can be easily incorporated into the nonlocal physics-informed deep learning framework: they are generated in discrete form during the preprocessing phase, and therefore they do not interfere with the deep-learning architectures, keeping their performance intact.

The outline of the paper is as follows. In Section~\ref{sec:PINN} we give a brief overview of the established (local) PINN framework, and its application to solid mechanics problems. In Section~\ref{sec:PINN_PDDO} we propose and describe an extension of the local (short-range) PINN to a nonlocal (long-range) PINN framework using PDDO. In Section~\ref{sec:example} we present the application of both local and nonlocal PINN to a representative example of elastoplastic deformation, corresponding to the indentation of a body by a rigid punch---an example that illustrates the effects of sharp gradients as a result of mixed displacement--traction boundary conditions. Finally, in Section~\ref{sec:conclusion} we discuss the results and summarize the main conclusions.

\section{Physics-Informed Deep Learning in Solid Mechanics}\label{sec:PINN}

In this section we provide a brief overview of the established (local) PINN framework \cite{Raissi2019}, and its application to forward modeling and parameter identification in solid mechanics, as described by elastoplasticity.

\subsection{Basics of the PINN framework}\label{ssec:PINN_basics}

In the PINN framework \cite{Raissi2019}, the solution space is constructed by a deep neural network with the independent variables (e.g., coordinates~$\mathbf{x}$) as the network inputs.  In this feed-forward network, each layer outputs data as inputs for the next layer through nested transformations.  Corresponding to the vector of input variables, $\mathbf{x}$, the output values, $f(\mathbf{x})$ can be mathematically expressed as
\begin{equation}\label{eq:pinn1}
    \mathbf{z}^\ell = \textrm{actf}(\mathbf{W}^{\ell-1}\mathbf{z}^{\ell-1} + \mathbf{b}^{\ell-1}) \quad \text{with} \quad \ell=1,2, \dots, L 
\end{equation}
where $\mathbf{z}^0 = \mathbf{x}$, ${z}^L={f}(\mathbf{x})$ and $\mathbf{z}^l$ represent the inputs, final outputs and hidden layer outputs of the network, and $\mathbf{W}^\ell$ and $\mathbf{b}^\ell$ represent the weights and biases of each layer, respectively. Note that lowercase and capital boldface letters are used to reflect vector and matrix components while scalars are shown with italic fonts. The activation function is denoted by $\textrm{actf}$; it renders the network nonlinear with respect to the inputs. 

The `trained' ${f}(\mathbf{x})$ can be considered as an approximate solution to the governing PDE.  It defines a mapping from inputs $\mathbf{x}$ to the field variable ${f}$ in the form of a multi-layer deep neural network, i.e., ${f}:~\mathbf{x} \mapsto \mathcal{N}(\mathbf{x};\mathbf{W},\mathbf{b})$, with $\mathbf{W}$ and $\mathbf{b}$ representing the set of all network parameters. The network inputs $\mathbf{x}$ can be temporal and spatial variables in reference to a Cartesian coordinate system, i.e., $\mathbf{x} = (x,y,t)$ in 2D.

In the PINN framework, the physics, described by a partial differential equation $\mathcal{P}({f})$ with $\mathcal{P}$ as the partial differential operator, is incorporated in the loss or cost function $\mathcal{L}$ along with the training data as
\begin{equation}\label{eq:pinn2}
    \mathcal{L} \equiv |{f} - {f}^*| + |\mathcal{P}(f) - 0^*|,
\end{equation}
where ${f}^*$ is the training dataset (which can be inside the domain or on the boundary), and $0^*$ represents the expected (true) value for the differential operation $\mathcal{P}({f})$ at any given training or sampling point. In all modern implementations of the deep-learning framework, such as Theano \cite{bergstra2010theano}, Tensorflow \cite{abadi2016tensorflow} and MXNet \cite{chen2015mxnet}, the partial derivatives in $\mathcal{P}$ can be performed using automatic differentiation (AD) \cite{baydin2017automatic}---a fundamental aspect of the PINN architecture. 
\begin{figure}[H] 
    \centering
    \includegraphics[width=1.0\linewidth]{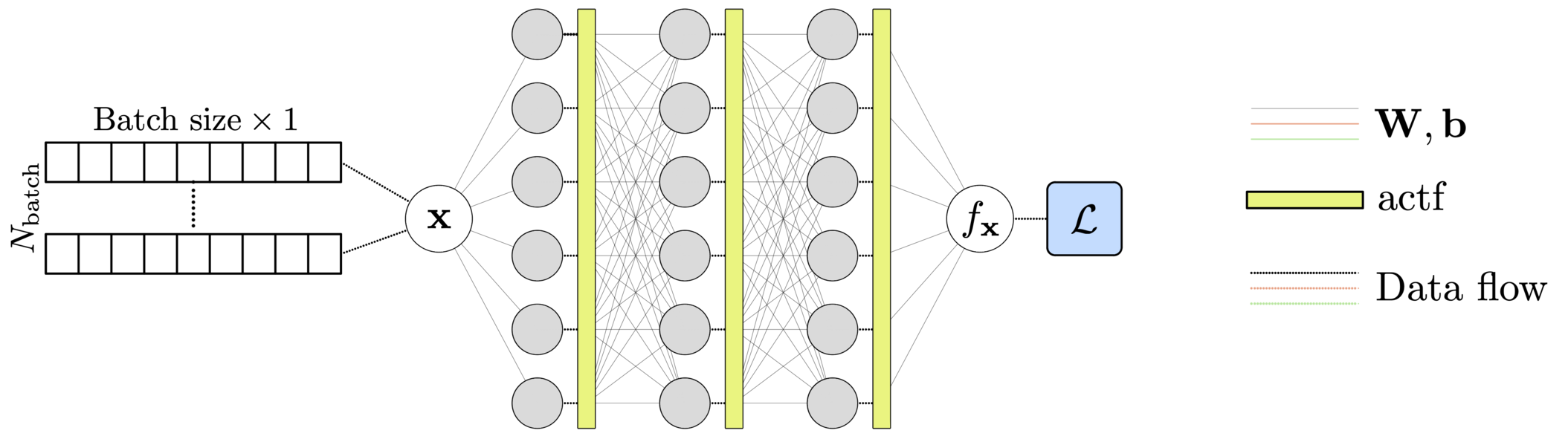}
    \caption{Local PINN architecture, defining the mapping ${f}:~\mathbf{x} \mapsto \mathcal{N}_f(\mathbf{x}; \mathbf{W}, \mathbf{b})$. }
    \label{fig:PINN-architecture}
\end{figure}
Different optimization algorithms exist for training a neural network; these include Adagrad \cite{duchi2011adaptive} and Adam \cite{kingma2014adam}. Several algorithmic parameters affect the rate of convergence of network training.  The algorithmic parameters include \textit{batch-size}, \textit{epochs}, \textit{shuffle} and \textit{patience}. Batch-size controls the number of samples from a dataset used to evaluate one gradient update. A batch-size of~1 is associated with a full stochastic gradient descent optimization. One epoch is one round of training on a dataset. If a dataset is shuffled, then a new round of training (epoch) results in an updated parameter set because the batch-gradients are evaluated on different batches. It is common to reshuffle a dataset many times and perform the back-propagation updates. 

The optimizer may, however, stop earlier if it finds that new rounds of epochs are not improving the loss function. This situation is described with the keyword \textit{patience}. It primarily occurs because the loss function is nonconvex. Therefore, the training needs to be tested with different starting points and in different directions to build confidence on the parameters evaluated from minimization of the loss function on a given dataset. Patience is the parameter that controls when the optimizer should stop the training. There are three basic strategies to train the network: (1)~generate a sufficiently large number of datasets and perform a one-epoch training on each dataset, (2)~work on one dataset over many epochs by reshuffling the data, and (3)~a combination of these. When dealing with synthetic data, all approaches are feasible.  In the original work on PINN \cite{Raissi2019}, the first strategy was used to train the model, with datasets being generated at random space locations at each epoch. This strategy, however, is often not applicable in real-world applications, especially when measurements are collected from sensors that are installed at fixed and limited spatial locations.

In this paper, we rely on SciANN \cite{haghighat2020sciann}, a recent implementation of PINN as a high-level Keras \cite{chollet2015keras} wrapper for physics-informed deep learning and scientific computations. Experimenting with all of the previously mentioned network choices can be easily done, with minimal coding, in SciANN \cite{haghighat2020sciann, haghighat2020deep}. 

\subsection{Solid mechanics with elastoplastic deformation}\label{ssec:elastoplas}

In the absence of body forces and neglecting inertia, the balance of linear momentum takes the form:
\begin{equation}\label{eq:equil}
    \sigma_{ij,j} = 0,
\end{equation}
for $i, j = x, y, z$, where $\sigma_{ij}$~is the Cauchy stress tensor, the subscript after a comma denotes differentiation, and a repeated subscript implies summation.

The linear elastic response of the material can be described by the stress--strain relations as
\begin{equation}\label{eq:elres}
    \sigma_{ij} = s_{ij} - p \delta_{ij},
\end{equation}
where the \emph{pressure} or volumetric stress is
\begin{equation}\label{eq:hp}
    p = -\sigma_{kk}/3 = -(\lambda + 2/3\mu) \varepsilon_{kk},
\end{equation}
and the deviatoric stress tensor is
\begin{equation}\label{eq:dp}
    s_{ij} = 2\mu e_{ij},
\end{equation}
in which
\begin{equation}\label{eq:eij1}
    e_{ij} = \varepsilon_{ij} - 1/3\varepsilon_{kk}\delta_{ij},
\end{equation}
with the strain tensor defined as
\begin{equation}\label{eq:stdisp}
   \varepsilon_{ij} = 1/2 (u_{i,j} + u_{j,i}),
\end{equation}
where $u_i$~are the components of the displacement field.

The nonlinear material response follows the classical description of elastoplastic behavior \cite{simo1998computational}. In particular, we adopt the von~Mises flow theory with a yield surface $\mathcal{F}$ defined as
\begin{equation}\label{eq:ps3c}
    \mathcal{F} = \sigma_{e} - (\sigma_{Y0} + H_p \bar{e}^p) \le 0,
\end{equation}
in which $\sigma_{Y0}$ is the initial yield stress, $H_p$ is the work hardening parameter, $\bar{e}^p$ is the equivalent plastic strain, and $\sigma_{e}$ is the effective stress. The plastic deformation occurs in the direction normal to the yield surface, i.e., $n_{ij}=\partial \mathcal{F}/\partial \sigma_{ij}$. We decompose the deviatoric strain tensor into its elastic and plastic components,
\begin{equation}\label{eq:eij2}
    e_{ij} = e_{ij}^e + e_{ij}^p.
\end{equation}
To account for plastic deformation, the equations describing the linear elastic material response, Eq.~\eqref{eq:dp} can be rewritten as
\begin{equation}\label{eq:dp2}
    s_{ij} = 2\mu (e_{ij} - e_{ij}^p),
\end{equation}
and
\begin{equation}\label{eq:eijp}
    e_{ij}^p = \left(\dfrac{3s_{ij}}{2\sigma_e}\right)\bar{e}^p.
\end{equation}
The effective stress~$\sigma_e$ is defined as
\begin{equation}\label{eq:es}
    \sigma_e = \sqrt{3J_2},
\end{equation}
with
\begin{equation}\label{eq:j2}
    J_2=\frac{1}{2}s_{ij}s_{ij}.
\end{equation}
The equivalent plastic strain, $e_{ij}^p$ can be obtained from Eq.~\eqref{eq:ps3c}, Eq.~\eqref{eq:dp2} and Eq.~\eqref{eq:eijp} as
\begin{equation}\label{eq:ps}
    \bar{e}^p = \dfrac{3\mu\bar{e}-\sigma_{Y0}}{3\mu+H_p}\ge 0,
\end{equation}
where
\begin{equation}\label{eq:ebar}
    \bar{e} = \sqrt{\dfrac{2}{3} e_{ij} e_{ij}}~.
\end{equation}

For linear elasticity under plane-strain conditions, the transverse component of strain, $\varepsilon_{zz}$, is identically equal to zero, and the transverse normal component of stress is evaluated as $\sigma_{zz}= \nu(\sigma_{xx}+ \sigma_{yy})$. For elastoplasticity, however, $\sigma_{zz}$ is not predefined while $\varepsilon_{zz}$ remains identically equal to zero.

\subsection{Local PINN for elastoplasticity}\label{ssec:localPINN}

Here we apply the PINN framework to the solution and inference of two-dimensional quasi-static mechanics. The input variables to the feed-forward neural network are the coordinates, $x$ and~$y$, and the output variables are the components of the displacement, $u_x$, $u_y$, strain tensor, $\varepsilon_{xx}$, $\varepsilon_{yy}$, $\varepsilon_{xy}$, and stress tensor, $\sigma_{xx}$, $\sigma_{yy}$, $\sigma_{xy}$. We define the loss function for \emph{linear elasticity} as:
\begin{equation}\label{eq:pinn3}
    \begin{split}
    \mathcal{L} &= |u_x-u_x^*|_{I_{u_x}} + |u_y-u^*_y|_{I_{u_y}} \\
                &+ |\sigma_{xx}-\sigma_{xx}^*|_{I_{\sigma_{xx}}} + |\sigma_{yy}-\sigma_{yy}^*|_{I_{\sigma_{yy}}} + |\sigma_{xy}-\sigma_{xy}^*|_{I_{\sigma_{xy}}} \\
                &+ |\varepsilon_{xx}-\varepsilon_{xx}^*|_{I_{\varepsilon_{xx}}} + |\varepsilon_{yy}-\varepsilon_{yy}^*|_{I_{\varepsilon_{yy}}} + |\varepsilon_{xy}-\varepsilon_{xy}^*|_{I_{\varepsilon_{xy}}} \\
                &+|\sigma_{xx,x} + \sigma_{xy,y} - 0^*|_I + |\sigma_{xy,x} + \sigma_{yy,y} - 0^*|_I \\
                &+ |-(\lambda+2/3\mu)(u_{x,x} + u_{y,y}) - p|_I \\
                &+ |2\mu e_{xx} - s_{xx}|_I + |2\mu e_{yy} - s_{yy}|_I + |2\mu e_{xy} - s_{xy}|_I
    \end{split}
\end{equation}
where the $\circ$ and $\circ^*$ components refer to predicted and true values, respectively. The set $I$ contains all sampling nodes.  The set $I_{\square}$ contains all sampling nodes for variable $\square$ where actual data exist. The terms in the loss function represent measures of the error in the displacement, strain and stress fields, the equilibrium equations, and the constitutive relations. 

Similarly, the loss function for \emph{elastoplasticity} is: 
\begin{equation}\label{eq:pinn4}
    \begin{split}
    \mathcal{L} &= |u_x-u_x^*|_{I_{u_x}} + |u_y-u^*_y|_{I_{u_y}} \\
        &+ |\sigma_{xx}-\sigma_{xx}^*|_{I_{\sigma_{xx}}} + |\sigma_{yy}-\sigma_{yy}^*|_{I_{\sigma_{yy}}} + |\sigma_{xy}-\sigma_{xy}^*|_{I_{\sigma_{xy}}} + |\sigma_{zz}-\sigma_{zz}^*|_{I_{\sigma_{zz}}} \\
        &+ |\varepsilon_{xx}-\varepsilon_{xx}^*|_{I_{\varepsilon_{xx}}} + |\varepsilon_{yy}-\varepsilon_{yy}^*|_{I_{\varepsilon_{yy}}} + |\varepsilon_{xy}-\varepsilon_{xy}^*|_{I_{\varepsilon_{xy}}} + |\varepsilon_{zz}-\varepsilon_{zz}^*|_{I_{\varepsilon_{zz}}} \\
        &+ |\sigma_{xx,x} + \sigma_{xy,y} - 0^*|_I + |\sigma_{xy,x} + \sigma_{yy,y} - 0^*|_I \\
        &+ |-(\lambda+2/3\mu)(u_{x,x} + u_{y,y})  - p|_I \\
        &+ |s_{xx} - 2\mu(e_{xx} - e_{xx}^p)|_I + |s_{yy} - 2\mu(e_{yy} - e_{yy}^p)|_I \\
        &+ |s_{zz} - 2\mu(e_{zz} - e_{zz}^p)|_I + |s_{xy} - 2\mu(e_{xy} - e_{xy}^p)|_I \\
        &+ |\bar{e}^p - \mathrm{ReLU}(\frac{3\mu \bar{e} - \sigma_{Y0}}{3\mu + H_p})|_I  \\
        &+ |e_{xx}^p - \frac{3}{2} \bar{e}^p \frac{s_{xx}}{\sigma_e}|_I + |e_{yy}^p - \frac{3}{2} \bar{e}^p \frac{s_{yy}}{\sigma_e}|_I \\
        &+ |e_{zz}^p - \frac{3}{2} \bar{e}^p \frac{s_{zz}}{\sigma_e}|_I + |e_{xy}^p - \frac{3}{2} \bar{e}^p \frac{s_{xy}}{\sigma_e}|_I
    \end{split}
\end{equation}
These loss functions are used for deep-learning-based solution of the governing PDEs as well as for identification of the model parameters.  The constitutive relations and governing equations are tested at all sampling (collocation) points, while data can be selectively imposed. The material parameters are treated as constant values in the network for the solution of governing PDEs.  However, they are treated as network parameters, which change during the training phase, during model identification (see Fig.~\ref{fig:PINN-architecture}). TensorFlow \cite{abadi2016tensorflow} permits such variables to be defined as Constant (PDE solution) or Variable (parameter identification) objects, respectively.

\section{Nonlocal PINN Architecture with the Peridynamics Differential Operator}\label{sec:PINN_PDDO}

Here we propose and describe an extension of the local (short-range) PINN with a single input~$\mathbf{x}$ to a nonlocal neural network that employs input variables in the form of family members $\mathcal{H}_{\mathbf{x}}$ of point $\mathbf{x}$, defined as $\mathcal{H}_{\mathbf{x}} = \{ \mathbf{x}' | w(\mathbf{x}' - \mathbf{x}) > 0 \}$. Each point $\mathbf{x}$ has its own unique family in its domain of interaction (an area in two-dimensional analysis). Given the relative position with reference to point~$\mathbf{x}$, $\boldsymbol{\xi}= \mathbf{x}-\mathbf{x}'$, the nondimensional weight function~$w(|\boldsymbol{\xi}|) = w(|\mathbf{x}-\mathbf{x}'|)$ represents the degree of interaction between the material points in each family. We define it as:
\begin{equation}\label{eq:pddo-pinn1}
    w({|\boldsymbol{\xi}|}) = e^{-4 |\boldsymbol{\xi}|^2 / \delta_{\mathbf{x}}^2},
\end{equation}
where the parameter $\delta_{\mathbf{x}}$, referred to as the horizon, defines the extent of the interaction domain (long-range interactions). In discrete form, the family members of point $\mathbf{x}$ are denoted as $\mathcal{H}_{\mathbf{x}} = (\mathbf{x}_{(1)}, \mathbf{x}_{(2)}, \dots, \mathbf{x}_{(N)})$, and their relative positions are defined as $\boldsymbol{\xi}_{(j)} = \mathbf{x} - \mathbf{x}_{(j)}$.
\begin{figure}[H] 
    \centering
    \includegraphics[width=0.5\linewidth]{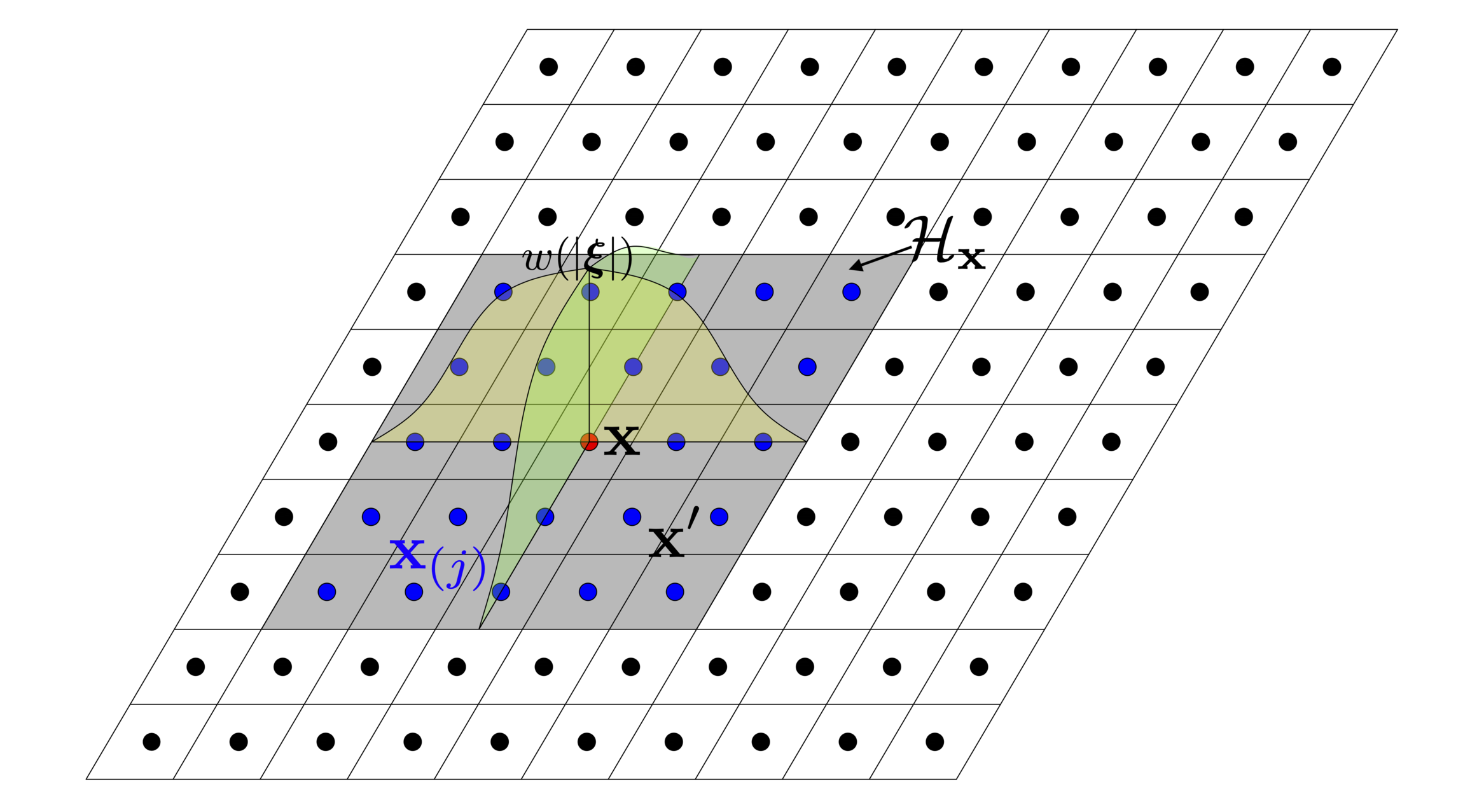}
    \caption{Interaction domain for point~$\mathbf{x}$, with $\mathbf{x}_{(j)}$ in its family.}
    \label{fig:PD_Interact}
\end{figure}
Silling \cite{silling2000reformulation} and Silling et al. \cite{silling2005meshfree} introduced the Peridynamic (PD) theory for failure initiation and growth in materials under complex loading conditions.  Recently, Madenci et al. \cite{madenci2016peridynamic, madenci2019peridynamic} introduced the Peridynamic Differential Operator (PDDO) to approximate the nonlocal representation of any function, such as a scalar field $f = f(\mathbf{x})$ and its derivatives at point $\mathbf{x}$, by accounting for the effect of its interactions with the other points, $\mathbf{x}_{(j)}$ in the domain of interaction $\mathcal{H}_{\mathbf{x}}$ (Fig.~\ref{fig:PD_Interact}).  

The derivation of PDDO employs Taylor Series Expansion (TSE), weighted integration and orthogonality (see \ref{sec:PDDO_derivation}). The major difference between PDDO and other existing local and nonlocal numerical differentiation methods is that PDDO leads to analytical expressions for arbitrary-order derivatives in integral form for a point with symmetric location in a circle. These analytical expressions, when substituted into the Navier displacement equilibrium equation, allow one to recover the PD equation of motion derived by Silling et al. \cite{silling2000reformulation}, which was based on the balance laws of continuum mechanics. The spatial integration can be performed numerically with simple quadrature techniques. As shown by Madenci et al. \cite{madenci2016peridynamic,madenci2019peridynamic,madenci2017numerical}, PDDO provides accurate evaluation of derivatives in the interior as well as the near the boundaries of the domain. 

The nonlocal PD representation of function $f(\mathbf{x})$ and its derivatives can be expressed in continuous and discrete forms as
\begin{equation}\label{eq:pddo-pinn2}
    f(\mathbf{x}) = \int_{\mathcal{H}_\mathbf{x}} f(\mathbf{x} + \boldsymbol{\xi})g_2^{00}(\boldsymbol{\xi})dA \approx \sum_{\mathbf{x}_{(j)} \in \mathcal{H}_\mathbf{x}} f_{(j)} g_{2~(j)}^{00} A_{(j)},
\end{equation}
\begin{equation}\label{eq:pddo-pinn3}
\begin{Bmatrix}
\dfrac{\partial f(\mathbf{x})}{\partial x}\\
\dfrac{\partial f(\mathbf{x})}{\partial y}
\end{Bmatrix}
=
\int_{\mathcal{H}_\mathbf{x}} f(\mathbf{x} + \boldsymbol{\xi})
\begin{Bmatrix}
g_2^{10}(\boldsymbol{\xi})\\
g_2^{01}(\boldsymbol{\xi})
\end{Bmatrix}
dA
\approx
\sum_{\mathbf{x}_{(j)}\in\mathcal{H}_\mathbf{x}} f_{(j)}
\begin{Bmatrix}
g^{10}_{2~(j)}\\
g^{01}_{2~(j)}
\end{Bmatrix}
A_{(j)},
\end{equation}
\begin{equation}\label{eq:pddo-pinn4}
\begin{Bmatrix}
\dfrac{\partial^2 f(\mathbf{x})}{\partial x^2}\\
\dfrac{\partial^2 f(\mathbf{x})}{\partial y^2}\\
\dfrac{\partial^2 f(\mathbf{x})}{\partial x \partial y}
\end{Bmatrix}
=
\int_{\mathcal{H}_\mathbf{x}} f(\mathbf{x} + \boldsymbol{\xi})
\begin{Bmatrix}
g_2^{20}(\boldsymbol{\xi})\\
g_2^{02}(\boldsymbol{\xi})\\
g_2^{11}(\boldsymbol{\xi})
\end{Bmatrix}
dA
\approx
\sum_{\mathbf{x}_{(j)} \in \mathcal{H}_\mathbf{x}} f_{(j)}
\begin{Bmatrix}
g^{20}_{2~(j)}\\
g^{02}_{2~(j)}\\
g^{11}_{2~(j)}
\end{Bmatrix}
A_{(j)},
\end{equation}
where $g_2^{p_1p_2}(\boldsymbol{\xi})$ with ($p,q = 0,1,2$) represent the PD functions obtained by enforcing the orthogonality condition of PDDO \cite{madenci2016peridynamic,madenci2019peridynamic}, and the integration is performed over the interaction domain.  The subscript $\circ_{(j)}$ reflects the discrete value of $f$, $g_2^{p_1, p_2}$, and~$A$ a family of point $\mathbf{x}_{(j)}$.

A \emph{nonlocal} neural network for a point~$\mathbf{x}$ and its family members can then be expressed as  
\begin{equation}\label{eq:pddo-pinn5}
    (f, f_{(1)}, \dots, f_{(N)}): (\mathbf{x}, \mathbf{x}_{(1)}, \dots, \mathbf{x}_{(N)}) \mapsto \tilde{\mathcal{N}}_f(\mathbf{x}, \mathbf{x}_{(1)}, \dots, \mathbf{x}_{(N)}; \mathbf{W}, \mathbf{b}).
\end{equation}
This network maps $\mathbf{x}$ and its family members $\mathcal{H}_{\mathbf{x}} = (\mathbf{x}_{(1)}, \dots, \mathbf{x}_{(N)})$ to the corresponding values of~$f$, i.e., $(f_{\mathbf{x}}, f_{\mathbf{x}_{(1)}}, \dots, f_{\mathbf{x}_{(N)}})$. With these output values, the nonlocal value of the field $f = f(\mathbf{x})$ and its derivatives can be readily constructed as 
\begin{equation}\label{eq:pddo-pinn6}
    \dfrac{\partial^{p_1} \partial^{p_2}}{\partial x^{p_1} \partial y^{p_2}} f(\mathbf{x}) = \tilde{\mathcal{N}}_f(\mathbf{x}, \mathbf{x}_{(1)}, \dots, \mathbf{x}_{(N)}; \mathbf{W}, \mathbf{b}) \cdot 
    \begin{Bmatrix}
        \mathcal{G}^{p_1p_2}_{2~\mathbf{x}} \\
        \mathcal{G}^{p_1p_2}_{2~\mathbf{x}_{(1)}} \\
        \dots \\
        \mathcal{G}^{p_1p_2}_{2~\mathbf{x}_{(N)}} \\
    \end{Bmatrix},
\end{equation}
where, $\mathcal{G}^{p_1p_2}_{2~\mathbf{x}_{(j)}} = g^{p_1p_2}_{2~\mathbf{x}_{(j)}} A_{\mathbf{x}_{(j)}}$. Here, the summation over discrete family points in Eq.~\eqref{eq:pddo-pinn2} is expressed as a dot product. Note that if $\delta_{\mathbf{x}}$ in the influence function \eqref{eq:pddo-pinn1} approaches zero, then $\mathcal{G}^{p_1p_2}_{2~\mathbf{x}} \rightarrow 1$ and $\mathcal{G}^{p_1p_2}_{2~\mathbf{x}_{(j)}} \rightarrow 0$, and we recover the local PINN architecture in Fig.~\ref{fig:PINN-architecture}.
\begin{figure}[H] 
    \centering
    \includegraphics[width=1.0\linewidth]{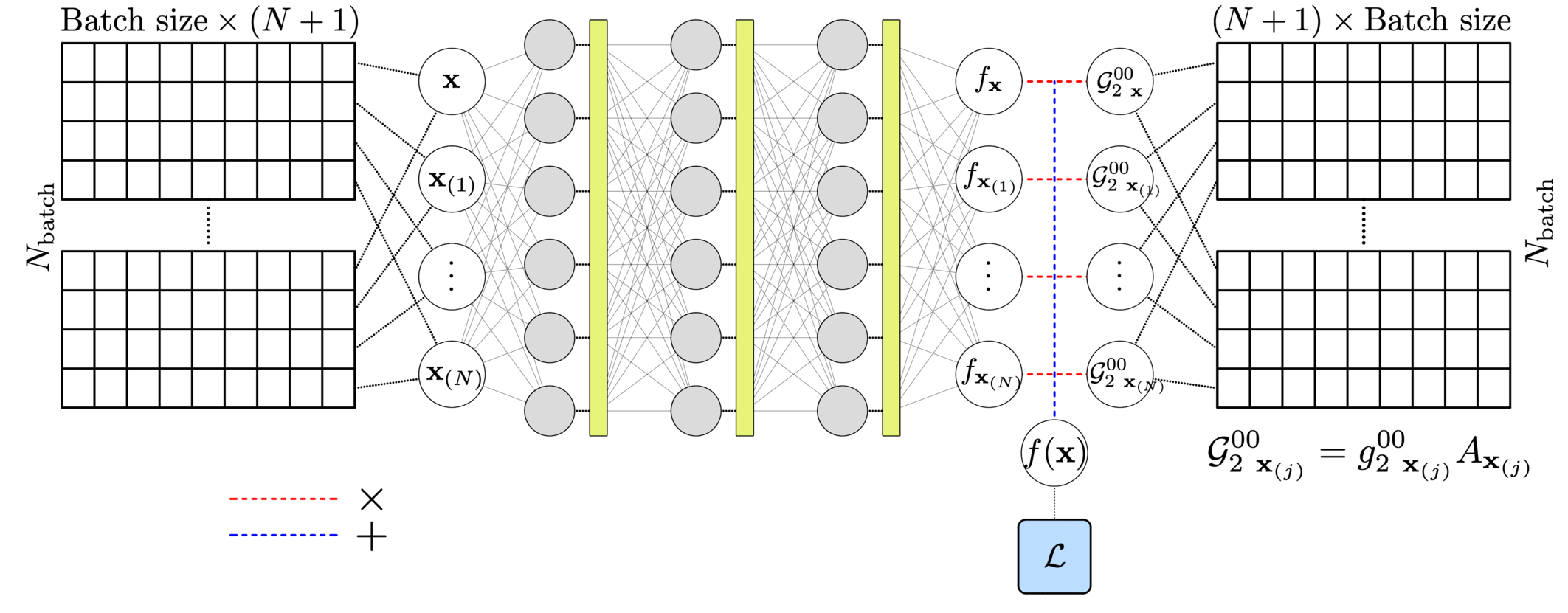}
    \caption{A nonlocal PDDO-PINN network architecture for approximation of function~$f(\mathbf{x})$.}
    \label{fig:PDONN-architecture}
\end{figure}

In our applications of PINN for solution and parameter inference, we make the distinction between two cases:
\begin{enumerate}
    \item We can use PDDO to approximate only the function, and use automatic differentiation (AD) to evaluate the derivatives. We refer to this case as AD-PDDO-PINN.
    \item We can instead use PDDO to approximate the function as well as its derivatives, instead of evaluating them from the network. We refer to this case as PDDO-PINN.
\end{enumerate}
As we will see, the use of PDDO-PINN enables the use of activation functions (such as $\mathrm{ReLU}$) and network architectures that cannot be used with local PINN---a capability that may lead to increased computational efficiency (per epoch) since it does not rely on extended graph computations associated with AD.

\section{Representative Example: Indentation of an Elastic or Elastoplastic Body} \label{sec:example}

In this section, we apply the different PINN formalisms to the solution and inference of a solid mechanics problem described by plane-strain elastoplastic deformation. The problem is designed to reflect a common scenario in boundary value problems: the presence of mixed boundary conditions, in which part of the boundary is subject to Dirichlet (displacement) boundary conditions, while part of the boundary is subject to Neumann (traction) boundary conditions. The sharp transition in type of boundary condition often leads to stress concentration (and sometimes a stress singularity). The problem we study here---indentation of a elastoplastic body---is subject to this stress concentration phenomenon, which poses a significant challenge to the application of existing deep learning techniques.

\subsection{Problem description}\label{ssec:problem_description}

\begin{figure}[H] 
    \centering
    \includegraphics[width=0.55\linewidth]{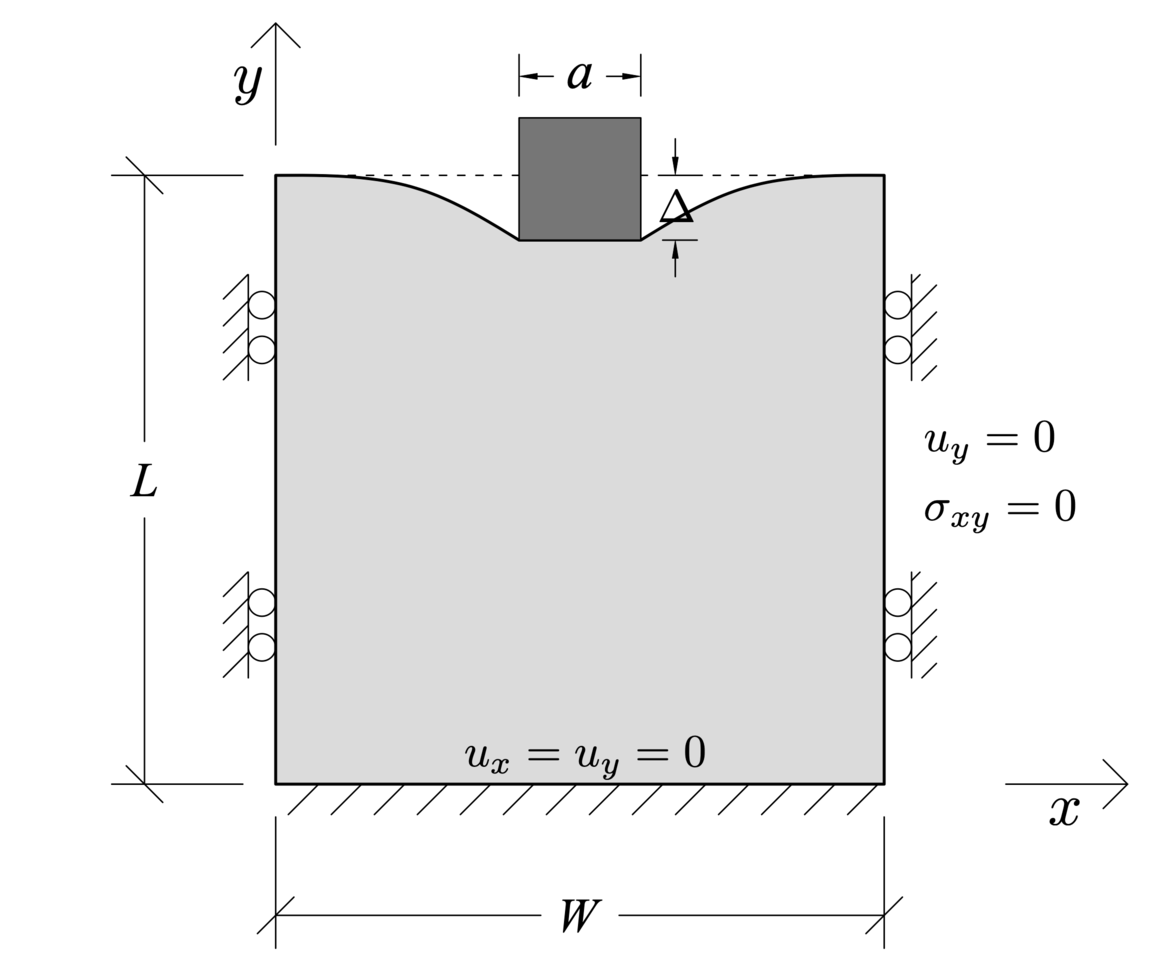}
    \caption{A square elastoplastic body under plane-strain conditions, and subject to a displacement~$\Delta$ in a portion of the top boundray via indentation by a rigid punch.}
    \label{fig:foundationM}
\end{figure}

We simulate the deformation of a square domain under plane-strain conditions, as a result of the indentation by a rigid punch (Fig.~\ref{fig:foundationM}). The body is constrained by roller support conditions along the lateral boundaries, and subject to fixed zero displacement along the bottom boundary. The dimensions of the domain are $W=L=1$~m, and thickness $h=1$~m. The width of the rigid punch is $a=0.2$~m, which indents the body at the top boundary a prescribed vertical displacement $\Delta=1$~mm. These boundary conditions can be expressed as
\begin{subequations}\label{eq:bcs}
\begin{align} 
    & u_x(x=0, y) = 0,        &&y \in (0, L), & \\
    & u_x(x=W, y) = 0,        &&y \in (0, L), & \\
    & u_x(x, y=0) = 0,        &&x \in (0, W), & \\
    & u_y(x, y=0) = 0,        &&x \in (0, W), & \\
    & u_y(x, y=L) = -\Delta,  &&x \in ((W-a)/2, (W+a)/2), & \\
    & \sigma_{yy}(x, y=L) =0, &&x \in (0, (W-a)/2), & \\
    & \sigma_{yy}(x, y=L) =0, &&x \in ((W+a)/2, W), & \\
    & \sigma_{xy}(x, y=L) =0, &&x \in (0, W).& 
\end{align}
\end{subequations}

The material exhibits elastic or elastic-plastic deformation with strain hardening. The elastic modulus, Poisson’s ratio, yield stress and hardening parameter of the material are specified as $E=70$~GPa, $\nu=0.3$, $\sigma_{Y0}=0.1$~GPa and $H_{0}=0.5$~GPa, respectively. The Lam\'e elastic constants, therefore, have values $\lambda=40.385$~GPa and $\mu=26.923$~GPa.

To generate synthetic (simulated) data to be used in the deep learning frameworks, we simulate the problem described above with the finite element method using COMSOL~\cite{COMSOL}.  The domain is discretized with a uniform mesh of size $100 \times 100$ elements of quartic Lagrange polynomials. 

\begin{figure}[H] 
    \centering
    \includegraphics[width=1.0\linewidth]{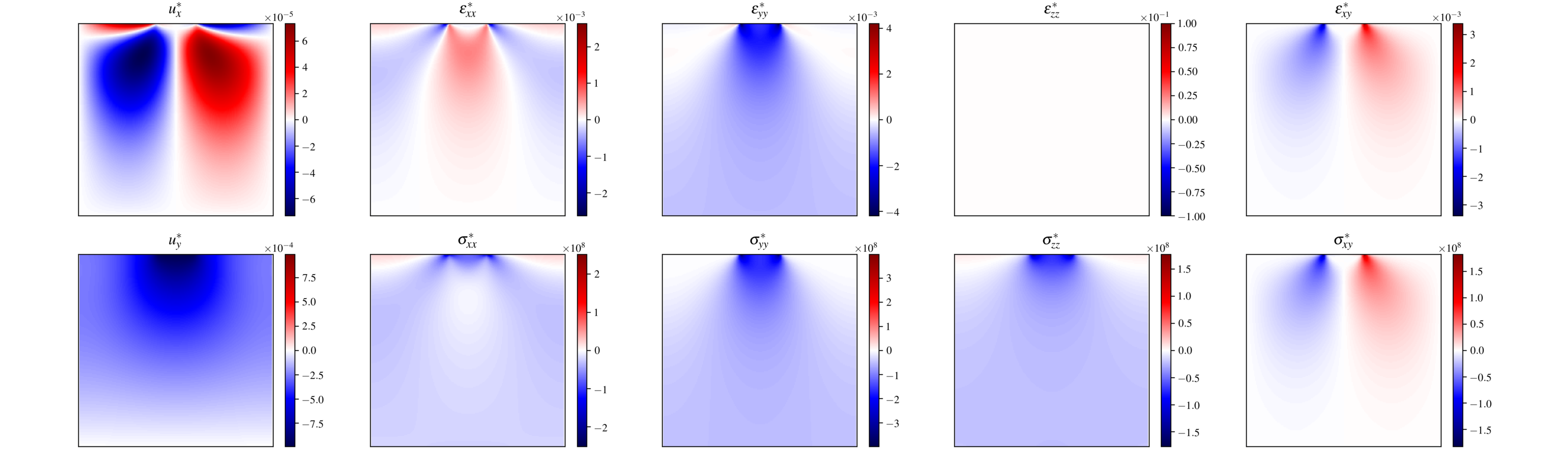}
    \caption{FEM reference solution for displacement, strain, and stress components in the case of purely linear elastic deformation.}
    \label{fig:fem_elasticity}
\end{figure}

\begin{figure}[H] 
    \centering
    \includegraphics[width=1.0\linewidth]{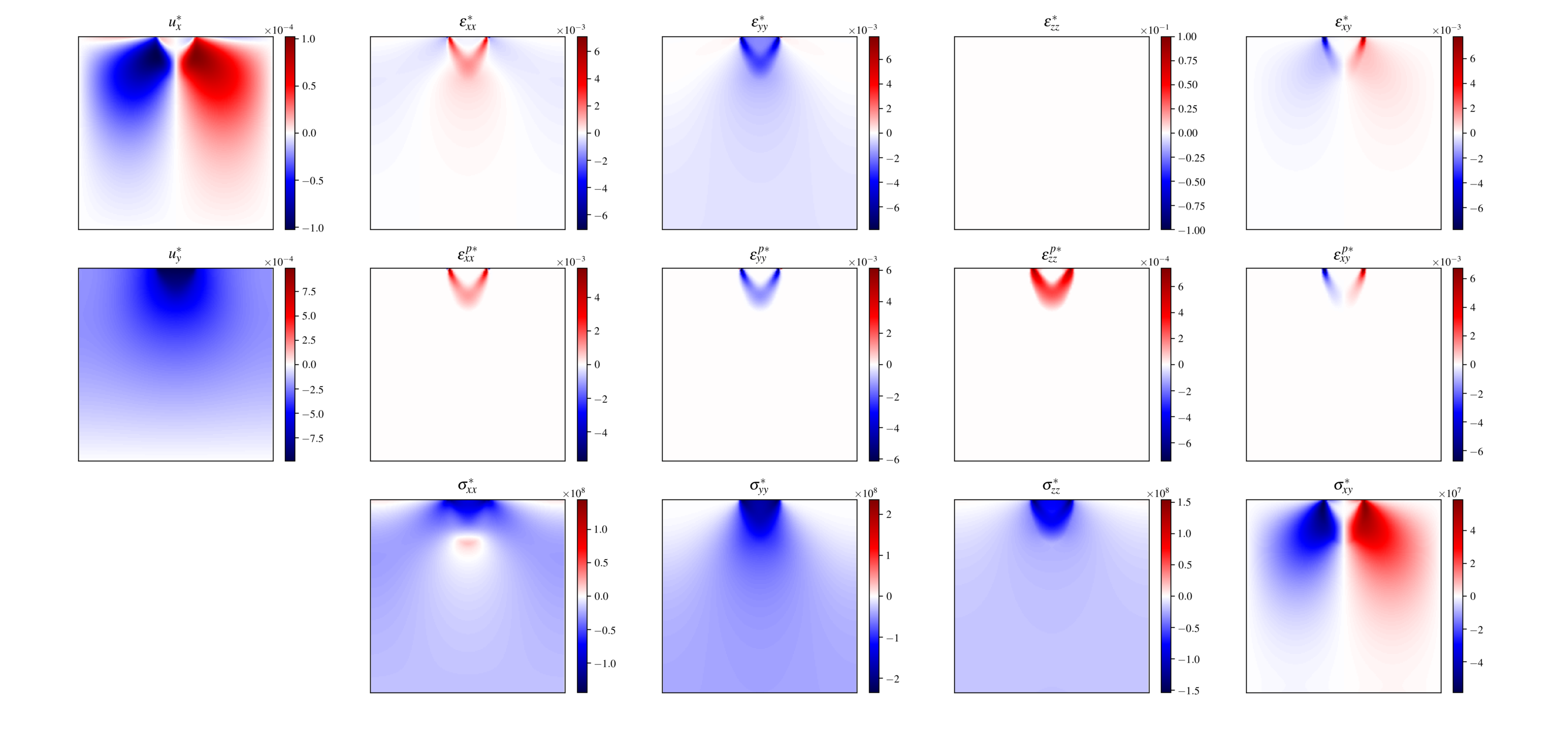}
    \caption{FEM reference solution for displacement, strain, and stress components in the case of elastic-plastic deformation.}
    \label{fig:fem_plasticity}
\end{figure}

The simulated displacement ($u_x$, $u_y$), strain ($\varepsilon_{xx}$, $\varepsilon_{yy}$, $\varepsilon_{zz}$, $\varepsilon_{xy}$) and stress ($\sigma_{xx}$, $\sigma_{yy}$, $\sigma_{zz}$, $\sigma_{xy}$) is computed for a purely linear elastic response (Fig.~\ref{fig:fem_elasticity}) and for elastic-plastic deformation (Fig.~\ref{fig:fem_plasticity}). It is apparent that the distribution of strain and stress components for the elastoplastic case are significantly different from those of the elastic case, with more localized deformation underneath the rigid punch. As expected, the plastic-strain components are zero in most of the domain, except in the vicinity of the corners of the punch, where it exhibits sharp gradients---a feature that, as we will see, poses a challenge for the approximation of the solution with a neural network.

\subsection{Local PINN results}\label{ssec:local_PINN_results}

We first apply the established (local) PINN framework for solution and parameter identification of the indentation problem described above \cite{Raissi2019, haghighat2020deep}. Training of the neural networks is performed with 10,000 training points (nodal solutions of the FEM solution). The convergence of the network training is sensitive to the choice of data-normalization and network size. After a trial-and-error approach, we selected the network architectures and parameters that led to the lowest value of the loss function and the highest accuracy of the physical model parameters. The selected feed-forward neural network has 4~hidden layers, each with 100~neuron units, and employs the hyperbolic-tangent activation function between layers.  We adopt batch-training with a total number of 20,000~epochs and a batch-size of~64. We use the Adam optimizer with a learning rate initialized to~0.0005 and decreased gradually to~0.000001 for the last epoch.

\begin{figure}[H] 
    \centering
    \includegraphics[width=1.0\linewidth]{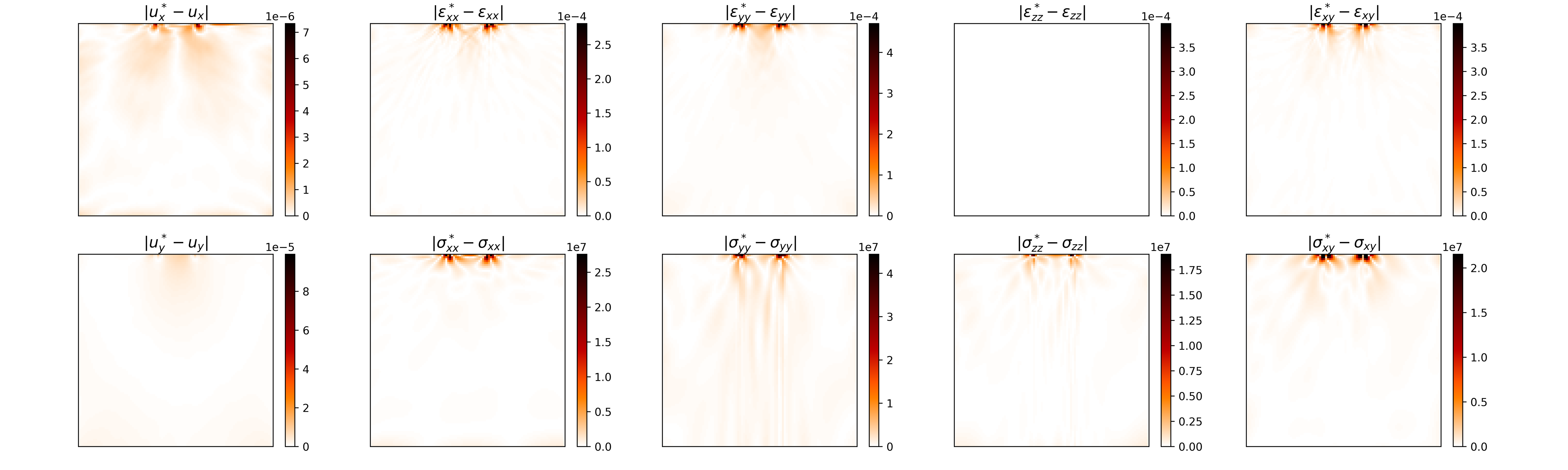}
    \caption{Difference between the local PINN predictions and the true data for displacement, strain, and stress components in the case of purely linear elastic deformation.}
    \label{fig:pinn_linear_fieldc}
\end{figure}

\begin{figure}[H] 
    \centering
    \adjincludegraphics[width=0.45\linewidth, trim={0 0 0 {0.5\height}}, clip]{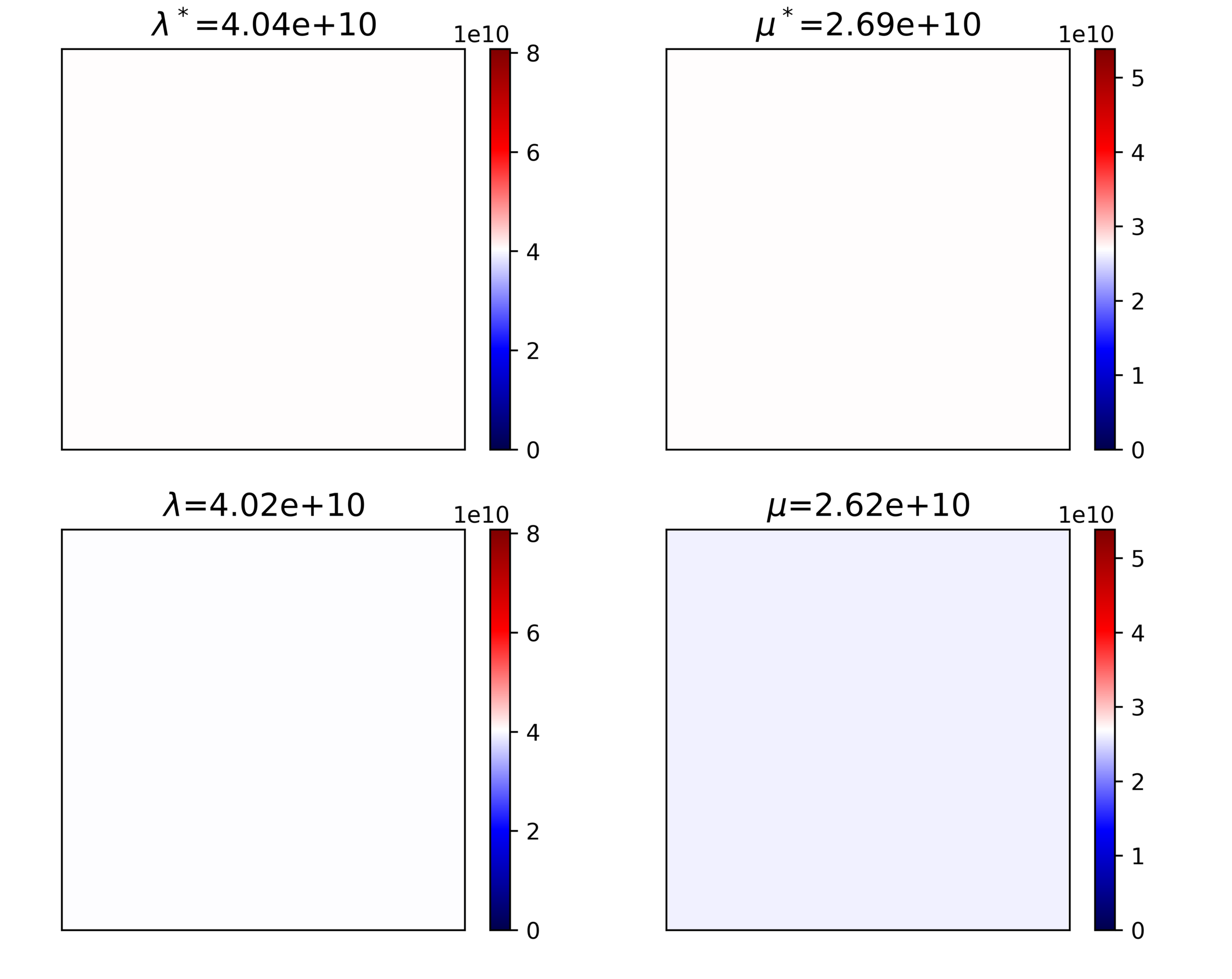}
    \caption{Local PINN predictions of the material parameters $\lambda$ and $\mu$ in the case of purely linear elastic deformation. White color indicates the true values of the parameters.)}
    \label{fig:pinn_linear_paramsc}
\end{figure}

The local PINN predictions for elastic deformation do capture the high-gradient regions near the corners of punch, but they are significantly diffused.  The differences between the local PINN predictions and the true data are shown in Fig.~\ref{fig:pinn_linear_fieldc}. The Lam\'e coefficients identified by the network are $\lambda=40.2$~GPa and $\mu=26.2$~GPa---an error of less than~3\% (Fig.~\ref{fig:pinn_linear_paramsc}).

\begin{figure}[H] 
    \centering
    \includegraphics[width=1.0\linewidth]{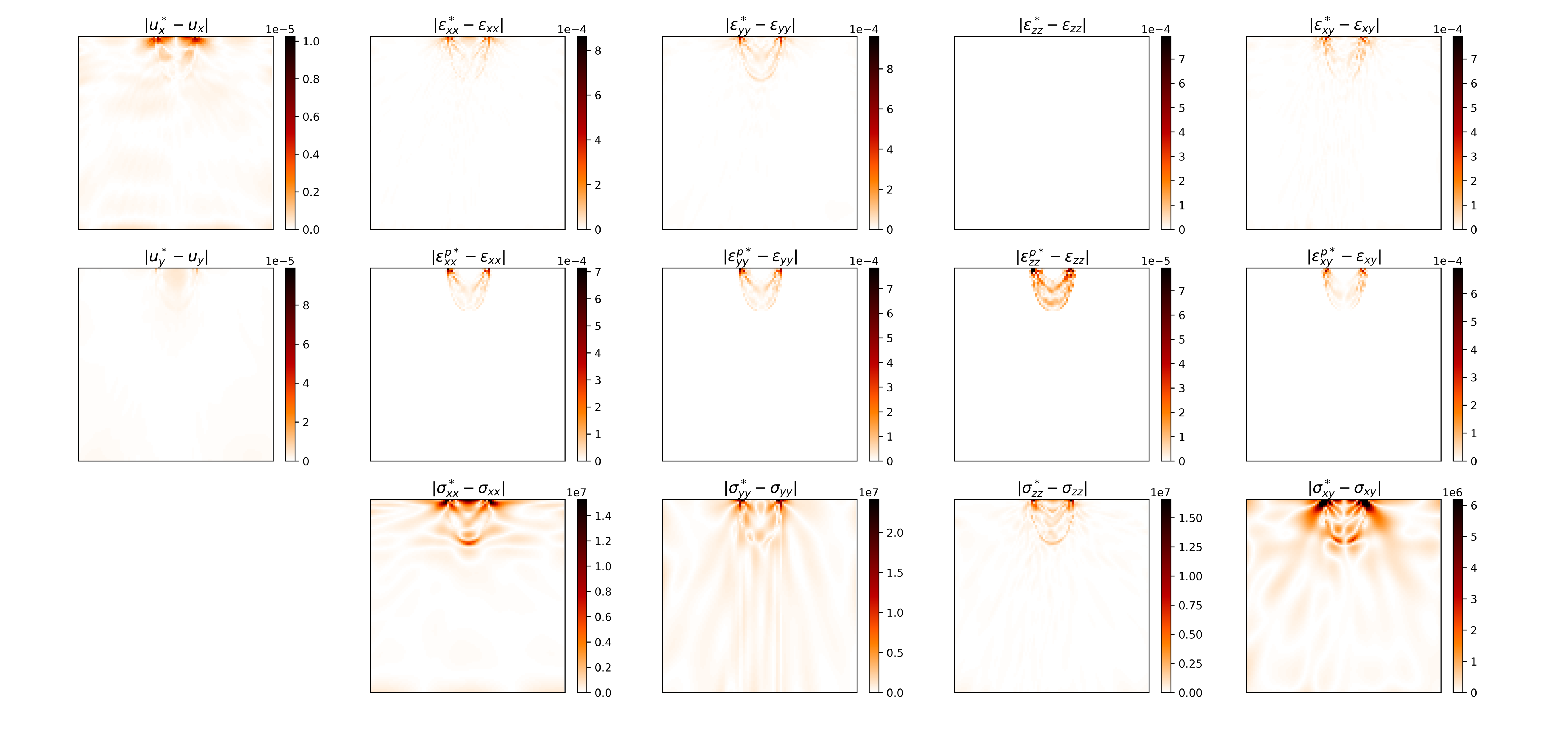}
    \caption{Difference between the local PINN predictions and the true data for displacement, strain, and stress components in the case of elastic-plastic deformation.}
    \label{fig:pinn_nonlinear_fieldc}
\end{figure}

\begin{figure}[H] 
    \centering
    \adjincludegraphics[width=1.0\linewidth, trim={0 0 0 {0.5\height}}, clip]{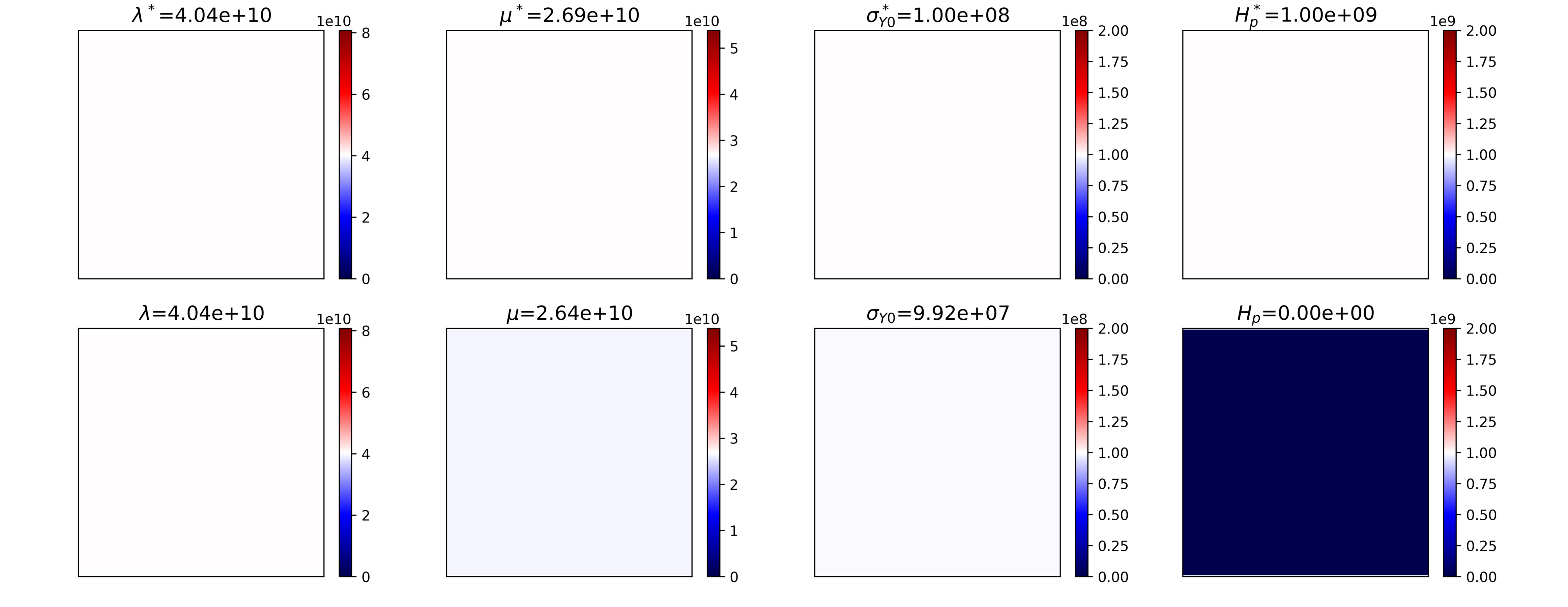}
    \caption{Local PINN predictions of the material parameters $\lambda$, $\mu$, $\sigma_{Y0}$ and $H_p$ in the case of elastic-plastic deformation. White color indicates the true values of the parameters.}
    \label{fig:pinn_nonlinear_paramsc}
\end{figure}

In the case of elastic-plastic deformation, $\sigma_{zz}$ depends on the plastic-strain components. Thus, the PINN architecture is defined with networks for $u_x$, $u_y$, $\sigma_{xx}$, $\sigma_{yy}$, $\sigma_{xy}$ and $\sigma_{zz}$. The error between the local PINN predictions and the true data are shown in Fig.~\ref{fig:pinn_nonlinear_fieldc}. In contrast with the local PINN predictions for the elastic case, the predictions for this case show poor quantitative agreement with the exact solution. The material parameters identified by the method are: $\lambda=40.4$~GPa, $\mu=26.4$~GPa, $\sigma_{Y0}=0.0992$~GPa and $H_p=0.00$~GPa. While the elastic Lam\'e coefficients and the yield stress are identified accurately, the method fails to identify the hardening parameter~$H_p$ (Fig.~\ref{fig:pinn_nonlinear_paramsc}). We speculate that this is due to the localized plastic deformation in narrow regions in the vicinity of the corners of the rigid punch (Fig.~\ref{fig:fem_plasticity}). Therefore, there are very few sampling points that contribute to the loss function with the
local PINN network. 

\subsection{Nonlocal PINN results}\label{ssec:nonlocal_PINN_results}

Given the relative success of the local PINN framework, but also the challenges faced in capturing the sharp gradients in the solution of the elastoplastic deformation problem, here we investigate the application of \emph{nonlocal} PINN to this problem.

The selected feed-forward neural networks are identical to those used in local PINN: they have 4~hidden layers, each with 100~neuron units, and with hyperbolic-tangent activation functions. ]In the construction of the PD functions, $g_2^{p_1p_2}(\boldsymbol{\xi})$, the TSE is truncated after the second-order derivatives ($N=2$, see ~\ref{sec:PDDO_derivation}).  The number of family members for each point depends on the order of approximation in the TSE; it is  $(N+1)$~points in each dimension, resulting in $(2N+3)\times(2N+3)$ for a square horizon in 2D~\cite{madenci2019peridynamic}.  Therefore, we choose a maximum number of 49~members as the nonlocal input features. Depending on the location of $\mathbf{x}$, the influence (degree of interaction) of some of these points (family members) might be zero. However, they are all incorporated in the construction of the nonlocal neural network to simplify the implementation procedure. 

In what follows we present the results for both AD-PDDO-PINN and PDDO-PINN architectures to the indentation problem with elastic-plastic deformation. 

\subsubsection{AD-PDDO-PINN}\label{sssec:ad-pddo-pinn}

The nonlocal deep neural network described by Eq.~\eqref{eq:pddo-pinn5} is employed to construct approximations for variables $u_x$, $u_y$, $\sigma_{xx}$, $\sigma_{yy}$ and $\sigma_{xy}$. They are evaluated as
\begin{equation}\label{eq:results1}
   f_\alpha(\mathbf{x}) = \tilde{\mathcal{N}}_\alpha (\mathbf{x}, \mathcal{H}_{\mathbf{x}}; \mathbf{W}, \mathbf{b}) \cdot
   \begin{Bmatrix}
    \mathcal{G}^{00}_{2~\mathbf{x}} \\
    \mathcal{G}^{00}_{2~\mathbf{x}_{(1)}} \\
    \dots \\
    \mathcal{G}^{00}_{2~\mathbf{x}_{(N)}} \\
   \end{Bmatrix},
\end{equation}
where $f_\alpha$ represents $u_x$, $u_y$, $\sigma_{xx}$, $\sigma_{yy}$, $\sigma_{xy}$. The derivatives are evaluated using automatic differentiation (AD). Since $f_{\alpha}(\mathbf{x})$ is a nonlocal function of $\mathbf{x}$ and its family points $\mathcal{H}_{\mathbf{x}}$, the differentiation of $f_{\alpha}$ is performed with respect to each family member using AD as
\begin{equation}\label{eq:results2}
   F^{p_1p_2}_\alpha(\mathbf{x}) = \dfrac{\partial^{p_1} \partial^{p_2}}{\partial x^{p_1} \partial y^{p_2}} f_\alpha(\mathbf{x}).
\end{equation}
In order to incorporate the effect of family members on the derivatives, the local AD differentiations are recast as 
\begin{equation}\label{eq:results2-2}
   \dfrac{\partial^{p_1} \partial^{p_2}}{\partial x^{p_1} \partial y^{p_2}} f_\alpha(\mathbf{x}) = \sum_{\mathbf{x}_{(j)}\in\mathcal{H}_{\mathbf{x}}} {F^{p_1p_2}_\alpha(\mathbf{x})} \mathcal{G}^{00}_{2~\mathbf{x}_{(j)}}. 
\end{equation}
%


%
%

\begin{figure}[H] 
    \centering
    \includegraphics[width=1.0\linewidth]{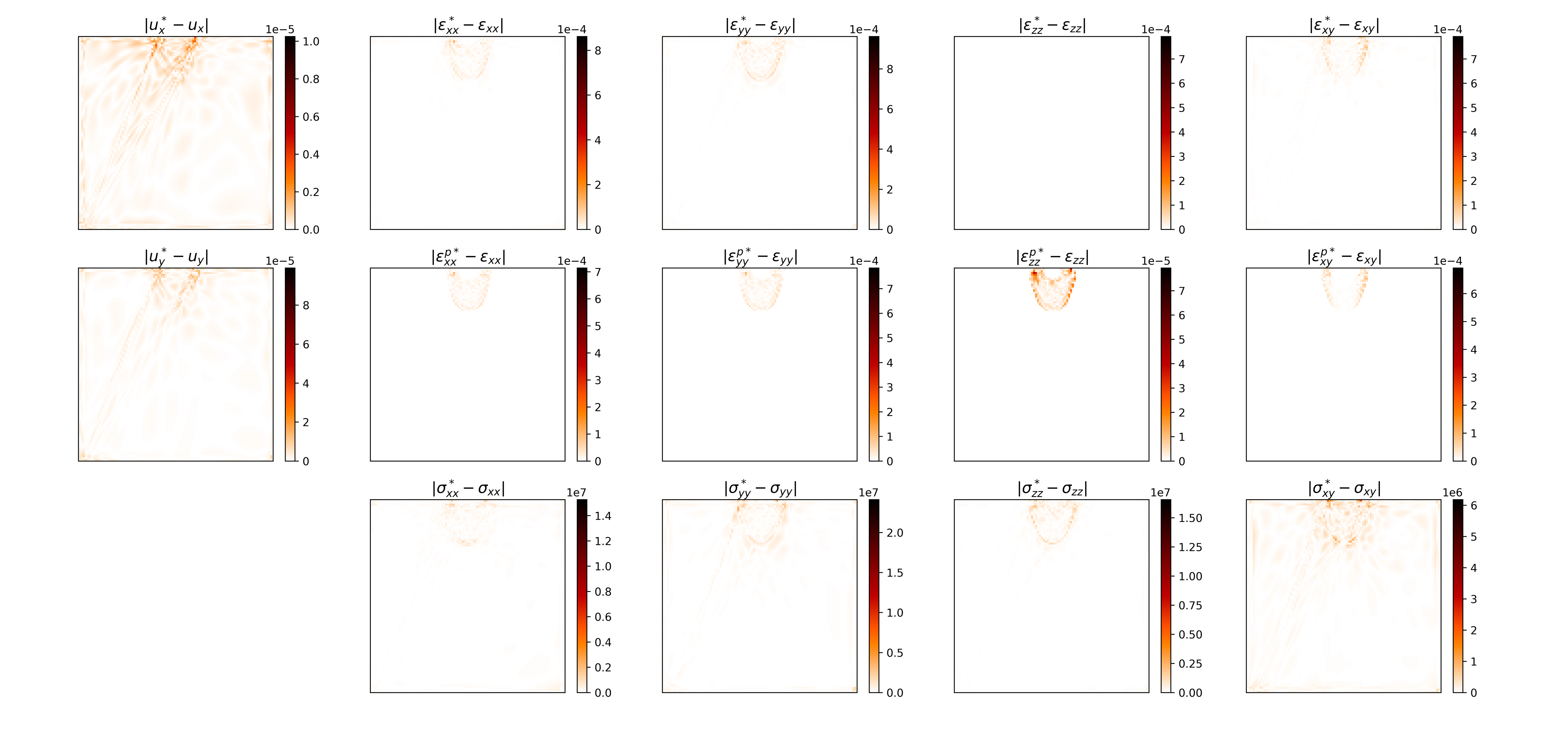}
    \caption{Difference between the nonlocal AD-PDDO-PINN predictions and the true data for displacement, strain, and stress components in the case of elastic-plastic deformation.}
    \label{fig:ad-pddo-pinn-nonlinear-field}
\end{figure}

\begin{figure}[H] 
    \centering
    \adjincludegraphics[width=1.0\linewidth, trim={0 0 0 {0.5\height}}, clip]{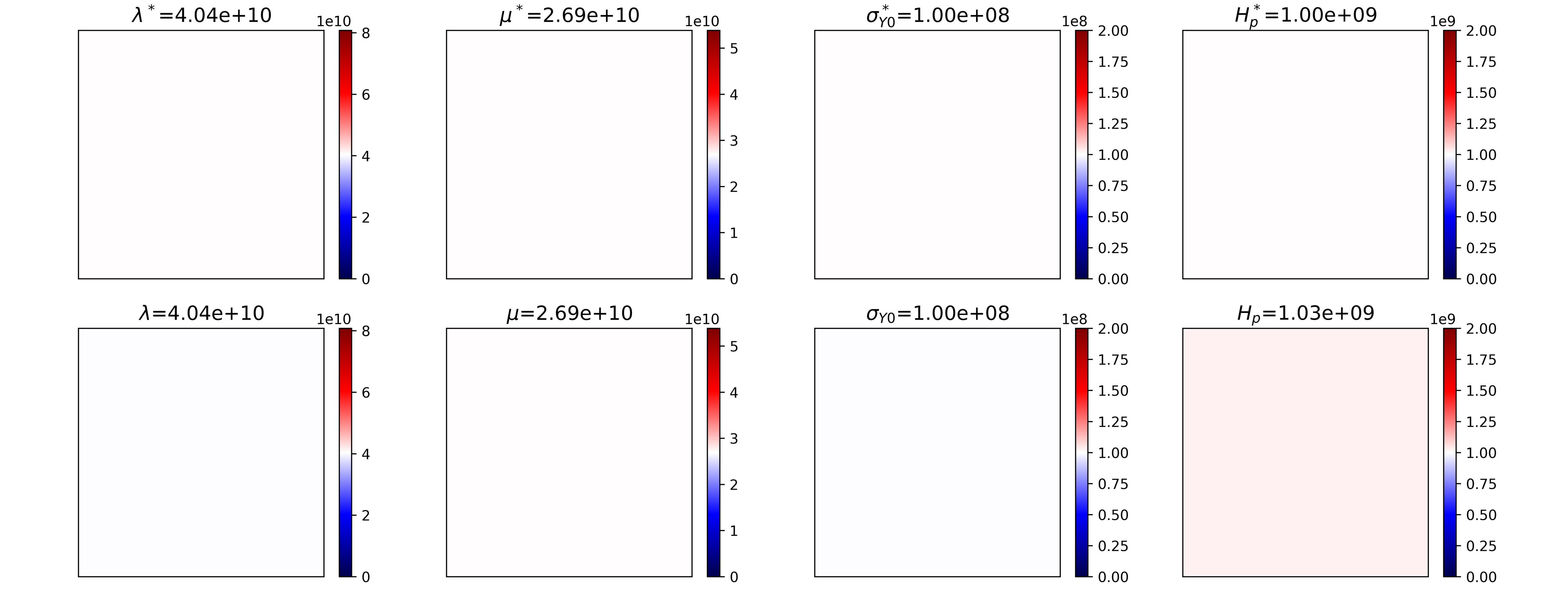}
    \caption{Nonlocal AD-PDDO-PINN predictions of the material parameters $\lambda$, $\mu$, $\sigma_{Y0}$ and $H_p$ in the case of elastic-plastic deformation. White color indicates the true values of the parameters.}
    \label{fig:ad-pddo-pinn-nonlinear-params}
\end{figure}

The differences between the AD-PDDO-PINN predictions and the true solution for the elastoplastic deformation case are shown in Fig.~\ref{fig:ad-pddo-pinn-nonlinear-field}. The value of the elastoplastic model parameters estimated by the method are: $\lambda=40.4$~GPa, $\mu=26.9$~GPa, $\sigma_{Y0}=0.10$~GPa and $H_p=1.03$~GPa (Fig.~\ref{fig:ad-pddo-pinn-nonlinear-params}). Both the solution and the model parameters are captured much more accurately than in the local PINN framework. In particular, the method reproduces the regions of high gradients in the solution, and is now able to accurately identify the hardening parameter~$H_p$.


\subsubsection{PDDO-PINN}\label{sssec:pddo-pinn}

We now employ the nonlocal deep neural network described by Eq.~\eqref{eq:pinn3} to construct approximations for variables $u_x$, $u_y$, $\sigma_{xx}$, $\sigma_{yy}$, $\sigma_{xy}$ \emph{and their derivatives}. These derivatives are evaluated as
\begin{equation}\label{eq:results3}
    \dfrac{\partial^{p_1} \partial^{p_2}}{\partial x^{p_1} \partial y^{p_2}} f_{\alpha}(\mathbf{x}) = \tilde{\mathcal{N}}_{\alpha}(\mathbf{x}, \mathcal{H}_{\mathbf{x}}; \mathbf{W}, \mathbf{b}) 
    \cdot
    \begin{Bmatrix}
     \mathcal{G}^{p_1p_2}_{2~\mathbf{x}} \\
     \mathcal{G}^{p_1p_2}_{2~\mathbf{x}_{(1)}} \\
     \dots \\
     \mathcal{G}^{p_1p_2}_{2~\mathbf{x}_{(N)}} \\
    \end{Bmatrix},
\end{equation}
where $f_\alpha$ represents $u_x$, $u_y$, $\sigma_{xx}$, $\sigma_{yy}$, $\sigma_{xy}$. 


%
%

\begin{figure}[H] 
    \centering
    \includegraphics[width=1.0\linewidth]{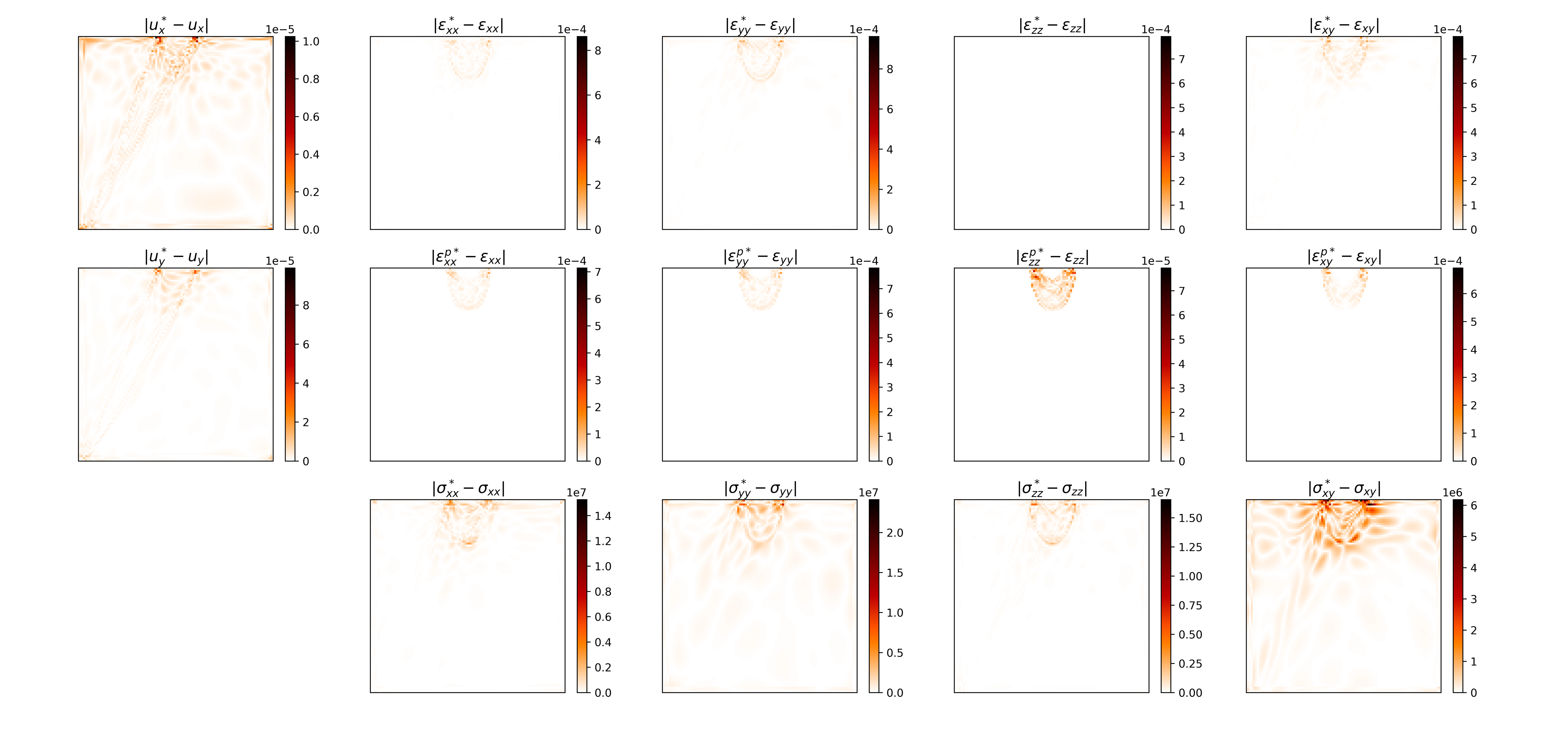}
    \caption{Difference between the nonlocal PDDO-PINN predictions and the true data for displacement, strain, and stress components in the case of elastic-plastic deformation.}
    \label{fig:pddo-pinn-nonlinear-field}
\end{figure}

\begin{figure}[H] 
    \centering
    \adjincludegraphics[width=1.0\linewidth, trim={0 0 0 {0.5\height}}, clip]{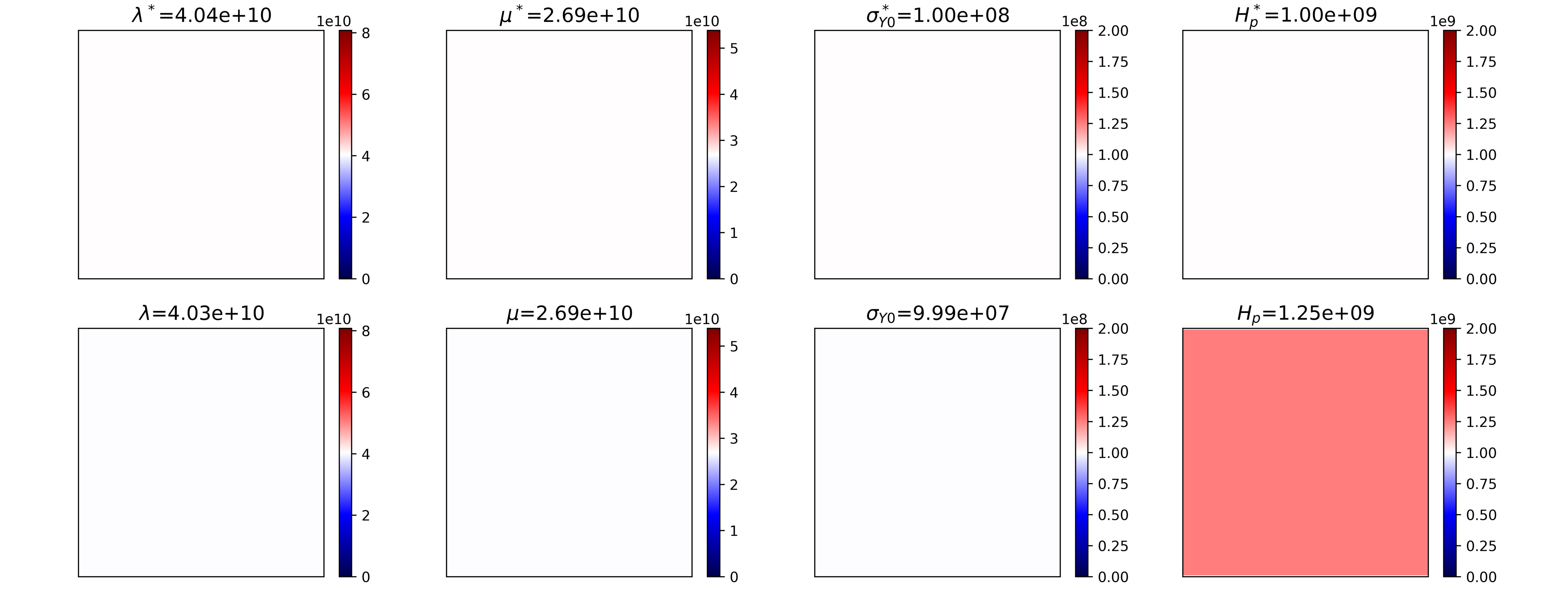}
    \caption{Nonlocal PDDO-PINN predictions of the material parameters $\lambda$, $\mu$, $\sigma_{Y0}$ and $H_p$ in the case of elastic-plastic deformation. White color indicates the true values of the parameters.}
    \label{fig:pddo-pinn-nonlinear-params}
\end{figure}

The errors in the PDDO-PINN solution for the elastoplastic deformation case are shown in Fig.~\ref{fig:pddo-pinn-nonlinear-field}, and the estimated elastoplastic model parameters are: $\lambda=40.3$~GPa, $\mu=26.9$~GPa, $\sigma_{Y0}=0.0999$~GPa and $H_p=1.25$~GPa (Fig.~\ref{fig:pddo-pinn-nonlinear-params}). The overall performance is better than that of local PINN, but less accurate than that of AD-PDDO-PINN. An advantage of the PDDO-PINN framework, however, is that it does not rely on automatic differentiation; therefore, the evaluation of derivatives through Eq.~\eqref{eq:results3} is faster for each epoch of training. 


\section{Discussion and Conclusions}\label{sec:conclusion}

\begin{figure}[H] 
    \centering
    \includegraphics[width=1.0\linewidth]{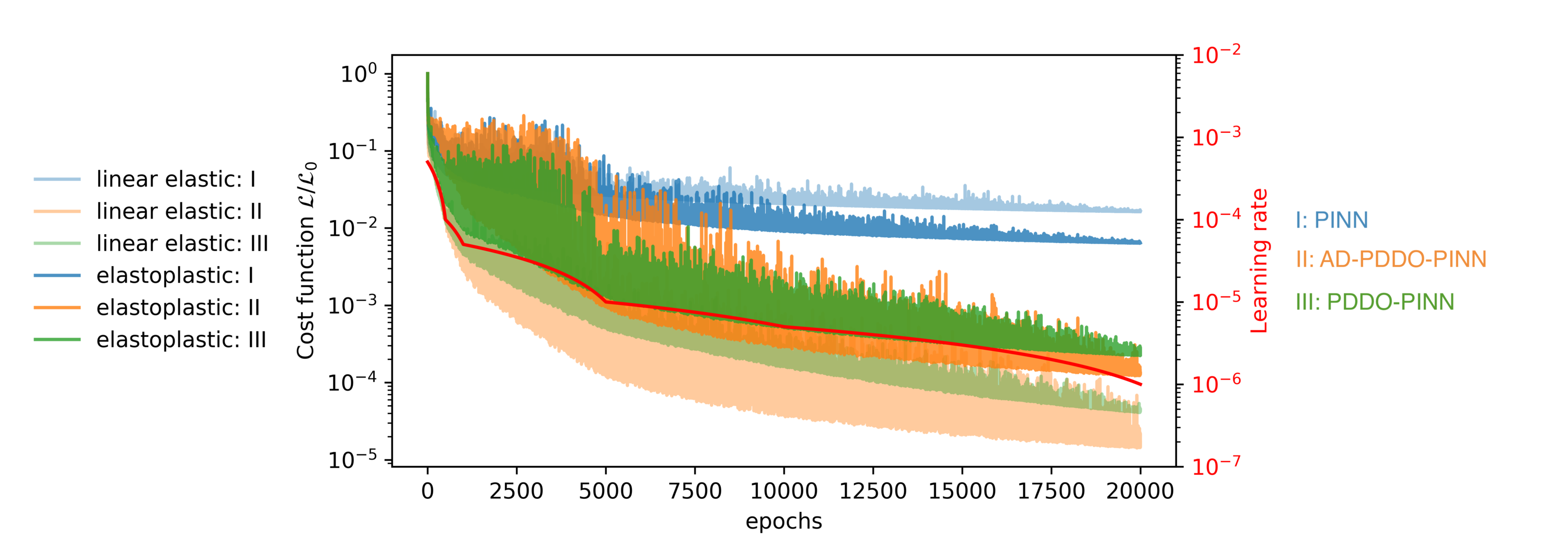}
    \caption{Convergence behavior of the different PINN frameworks (I:~local PINN, II:~nonlocal AD-PDDO-PINN, and III:~nonlocal PDDO-PINN), showing the evolution of the normalized loss function~$\mathcal{L}$ (left axis) and the learning rate (right axis) as a function of the number of epochs for both the linear-elastic and nonlinear elastoplastic deformation cases.}
    \label{fig:losses_summary}
\end{figure}

The results of the previous section demonstrate the benefits of the nonlocal PINN framework in the reconstruction of the deformation and parameter identification for solid-mechanics problems with sharp gradients in the solution, compared with those obtained with the local PINN architecture. This improved performance is also apparent from examination of the evolution of the normalized loss function~$\mathcal{L}$ for the different architectures (Fig.~\ref{fig:losses_summary}), illustrating the faster convergence and lower final value of~$\mathcal{L}/\mathcal{L}_0$ of the nonlocal PINN approaches. 

In summary, we have introduced a \emph{nonlocal} approach to Physics-Informed Neural Networks (PINN) using the Peridynamic Differential Operator (PDDO). In the limit when the interaction range~$\delta_{\mathbf{x}}$ approaches zero, the method reverts to the local PINN model. We have presented two versions of the proposed approach: one with automatic differentiation using the neural network (AD-PDDO-PINN), and the other with analytical evaluation of the derivatives relying on PDDO functions (PDDO-PINN). The PD functions can be readily and efficiently incorporated in the neural network architecture and, therefore, the nonlocality does not degrade the performance of modern deep-learning algorithms.  We have applied both versions of nonlocal PINN to the solution and identification of material parameters in solid mechanics. Specifically, we focused on the solution and inference of linear-elastic and elastoplastic deformation in a domain subjected to indentation by a rigid punch. The resulting boundary value problem is challenging because of the mixed displacement--traction boundary conditions along the top boundary, which result in localized deformation and sharp gradients in the solution. We have shown that the PDDO framework is able to capture the stress and strain concentrations with global functions and, as a result, leads to the superior behavior of nonlocal PINN both in terms of the accuracy of the solution and the estimated model parameters. While many questions remain with regard to the selection of network size, order of the PDDO approximation and training optimization algorithms, these results suggest that nonlocal PINN may offer a powerful framework for simulation and discovery of partial differential equations whose solution develops sharp gradients.

\section*{Acknowledgments}
RJ and EH conducted this work as a part of KFUPM-MIT collaborative agreement `Multiscale Reservoir Science'. EM and ACB performed this work as part of the ongoing research at the MURI Center for Material Failure Prediction through Peridynamics at the University of Arizona (AFOSR Grant No. FA9550-14-1-0073).

\appendix
\section{PDDO Derivation}\label{sec:PDDO_derivation}

According to the 2nd-order TSE in a 2-dimensional space, the following expression holds
\begin{equation}
f(\mathbf{x}+\boldsymbol{\xi}) = f(\mathbf{x}) + \xi_1 \dfrac{\partial f(\mathbf{x})}{\partial x_1} + \xi_2 \dfrac{\partial f(\mathbf{x})}{\partial x_2} + \dfrac{1}{2!} \xi_1^2 \dfrac{\partial^2 f(\mathbf{x})}{\partial x_1^2} + \dfrac{1}{2!} \xi_2^2 \dfrac{\partial^2 f(\mathbf{x})}{\partial x_2^2} + \xi_1 \xi_2 \dfrac{\partial^2 f(\mathbf{x})}{\partial x_1 \partial x_2} + \mathcal{R},
\end{equation}
where $\mathcal{R}$ is the remainder. Multiplying each term with PD functions, $g_2^{p_1p_2}(\boldsymbol{\xi})$ and integrating over the domain of interaction (family), $\mathcal{H}_{\mathbf{x}}$, results in
\begin{equation}
\begin{split}
\int_{\mathcal{H}_{\mathbf{x}}}f(\mathbf{x}+\boldsymbol{\xi})g_2^{p_1p_2}(\boldsymbol{\xi})dV &= f(\mathbf{x})\int_{\mathcal{H}_{\mathbf{x}}}g_2^{p_1p_2}(\boldsymbol{\xi})dV +\dfrac{\partial f(\mathbf{x})}{\partial x_1}\int_{\mathcal{H}_{\mathbf{x}}}\xi_1 g_2^{p_1p_2}(\boldsymbol{\xi})dV\\
&~+\dfrac{\partial f(\mathbf{x})}{\partial x_2}\int_{\mathcal{H}_{\mathbf{x}}}\xi_2 g_2^{p_1p_2}(\boldsymbol{\xi})dV +
  \dfrac{\partial^2 f(\mathbf{x})}{\partial x_1^2}\int_{\mathcal{H}_{\mathbf{x}}}\dfrac{1}{2!} \xi_1^2g_2^{p_1p_2}(\boldsymbol{\xi})dV\\
&~+\dfrac{\partial^2 f(\mathbf{x})}{\partial x_2^2}\int_{\mathcal{H}_{\mathbf{x}}}\dfrac{1}{2!} \xi_2^2 g_2^{p_1p_2}(\boldsymbol{\xi})dV +
\dfrac{\partial^2 f(\mathbf{x})}{\partial x_1 \partial x_2}\int_{\mathcal{H}_{\mathbf{x}}}\xi_1 \xi_2g_2^{p_1p_2}(\boldsymbol{\xi})dV,
\end{split}
\end{equation}
in which the point $\mathbf{x}$ is not necessarily symmetrically located in the domain of interaction. The initial relative position, $\boldsymbol{\xi}$, between the material points $\mathbf{x}$ and $\mathbf{x}'$ can be expressed as $\boldsymbol{\xi} = \mathbf{x} - \mathbf{x}'$. This ability permits each point to have its own unique family with an arbitrary position.  Therefore, the size and shape of each family can be different, and they significantly influence the degree of nonlocality.  The degree of interaction between the material points in each family is specified by a nondimensional weight function, $w(|\boldsymbol{\xi}|)$, which can vary from point to point. The interactions become more local with decreasing family size. Thus, the family size and shape are important parameters.  In general, the family of a point can be nonsymmetric due to nonuniform spatial discretization. Each point has its own family members in the domain of interaction (family), and occupies an infinitesimally small entity such as volume, area or a distance.

The PD functions are constructed such that they are orthogonal to each term in the TSE as
\begin{equation}
\dfrac{1}{n_1!n_2!}\int_{\mathcal{H}_{\mathbf{x}}}\xi_1^{n_1}\xi_2^{n_2}g_2^{p_1p_2}(\boldsymbol{\xi})dV = \delta_{n_1p_1}\delta_{n_2p_2},
\end{equation}
with ($n_1,n_2,p,q=0,1,2$) and $\delta$ is the Kronecker symbol. Enforcing the orthogonality conditions in the TSE leads to the nonlocal PD representation of the function itself and its derivatives as
\begin{subequations}
\begin{align}
f(\mathbf{x}) &= \int_{\mathcal{H}_{\mathbf{x}}}f(\mathbf{x} + \boldsymbol{\xi})g_2^{00}(\boldsymbol{\xi})dV,\\
\begin{Bmatrix}
\dfrac{\partial f(\mathbf{x})}{\partial x}\\
\dfrac{\partial f(\mathbf{x})}{\partial y}
\end{Bmatrix}
&=
\int_{\mathcal{H}_{\mathbf{x}}}f(\mathbf{x} + \boldsymbol{\xi})
\begin{Bmatrix}
g_2^{10}(\boldsymbol{\xi})\\
g_2^{01}(\boldsymbol{\xi})
\end{Bmatrix}
dV,\\
\begin{Bmatrix}
\dfrac{\partial^2 f(\mathbf{x})}{\partial x^2}\\
\dfrac{\partial^2 f(\mathbf{x})}{\partial y^2}\\
\dfrac{\partial^2 f(\mathbf{x})}{\partial x \partial y}
\end{Bmatrix}
&=
\int_{\mathcal{H}_{\mathbf{x}}}f(\mathbf{x} + \boldsymbol{\xi})
\begin{Bmatrix}
g_2^{20}(\boldsymbol{\xi})\\
g_2^{02}(\boldsymbol{\xi})\\
g_2^{11}(\boldsymbol{\xi})
\end{Bmatrix}
dV.
\end{align}
\end{subequations}

The PD functions can be constructed as a linear combination of polynomial basis functions
\begin{equation}
\begin{split}
g_2^{p_1p_2} =& a_{00}^{p_1p_2}w_{00}(|\mathbf{\boldsymbol{\xi}}|) + a_{10}^{p_1p_2}w_{10}(|\mathbf{\boldsymbol{\xi}}|)\xi_1 + a_{01}^{p_1p_2}w_{01}(|\mathbf{\boldsymbol{\xi}}|)\xi_2 + a_{20}^{p_1p_2}w_{20}(|\mathbf{\boldsymbol{\xi}}|)\xi_1^2\\
&~+a_{02}^{p_1p_2}w_{02}(|\mathbf{\boldsymbol{\xi}}|)\xi_2^2 + a_{11}^{p_1p_2}w_{11}(|\mathbf{\boldsymbol{\xi}}|)\xi_1\xi_2,
\end{split}
\end{equation}
where $a_{q_1q_2}^{p_1p_2}$ are the unknown coefficients, $w_{q_1q_2}(|\mathbf{\boldsymbol{\xi}}|)$ are the influence functions, and $\xi_1$ and $\xi_2$ are the components of the vector $\mathbf{\boldsymbol{\xi}}$. Assuming $w_{q_1q_2}(|\mathbf{\boldsymbol{\xi}}|) = w(|\mathbf{\boldsymbol{\xi}}|)$ and incorporating the PD functions into the orthogonality equation lead to a system of algebraic equations for the determination of the coefficients as
\begin{equation}
\mathbf{A}\mathbf{a}=\mathbf{b},
\end{equation}
where
\begin{subequations}
\begin{align}
\mathbf{A} &= \int_{\mathcal{H}_{\mathbf{x}}}w(|\mathbf{\boldsymbol{\xi}}|)
\begin{bmatrix}
1 & \xi_1 & \xi_2 & \xi_1^2 & \xi_2^2 & \xi_1\xi_2\\
\xi_1 & \xi_1^2 & \xi_1\xi_2 & \xi_1^3 & \xi_1\xi_2^2 & \xi_1^2\xi_2\\
\xi_2 & \xi_1\xi_2 & \xi_2^2 & \xi_1^2\xi_2 & \xi_2^3 & \xi_1\xi_2^2\\
\xi_1^2 & \xi_1^3 & \xi_1^2\xi_2 & \xi_1^4 & \xi_1^2\xi_2^2 & \xi_1^3\xi_2\\
\xi_2^2 & \xi_1\xi_2^2 & \xi_2^3 & \xi_1^2\xi_2^2 & \xi_2^4 & \xi_1\xi_2^3\\
\xi_1\xi_2 & \xi_1^2\xi_2 & \xi_1\xi_2^2 & \xi_1^3\xi_2 & \xi_1\xi_2^3 & \xi_1^2\xi_2^2\\
\end{bmatrix}dV,\\
\mathbf{a} &= 
\begin{bmatrix}
a_{00}^{00} & a_{10}^{00} & a_{01}^{00} & a_{20}^{00} & a_{02}^{00} & a_{11}^{00}\\
a_{00}^{10} & a_{10}^{10} & a_{01}^{10} & a_{20}^{10} & a_{02}^{10} & a_{11}^{10}\\
a_{00}^{01} & a_{10}^{01} & a_{01}^{01} & a_{20}^{01} & a_{02}^{01} & a_{11}^{01}\\
a_{00}^{20} & a_{10}^{20} & a_{01}^{20} & a_{20}^{20} & a_{02}^{20} & a_{11}^{20}\\
a_{00}^{02} & a_{10}^{02} & a_{01}^{02} & a_{20}^{02} & a_{02}^{02} & a_{11}^{02}\\
a_{00}^{11} & a_{10}^{11} & a_{01}^{11} & a_{20}^{11} & a_{02}^{11} & a_{11}^{11}\\
\end{bmatrix},\\
\mathbf{b} &= 
\begin{bmatrix}
1 & 0 & 0 & 0 & 0 & 0\\
0 & 1 & 0 & 0 & 0 & 0\\
0 & 0 & 1 & 0 & 0 & 0\\
0 & 0 & 0 & 2 & 0 & 0\\
0 & 0 & 0 & 0 & 2 & 0\\
0 & 0 & 0 & 0 & 0 & 1\\
\end{bmatrix}.
\end{align}
\end{subequations}

After determining the coefficients $a_{q_1q_2}^{p_1p_2}$ via $\mathbf{a}=\mathbf{A}^{-1}\mathbf{b}$, the PD functions $g_2^{p_1p_2}(\boldsymbol{\xi})$ can be constructed. The detailed derivations and the associated computer programs can be found in \cite{madenci2019peridynamic}. The PDDO is nonlocal; however, in the limit as the horizon size approaches zero, it recovers the local differentiation as proven by Silling and Lehoucq \cite{silling2008convergence}.

\bibliographystyle{elsarticle-num} 
\bibliography{refs}

\end{document}